%% file: revision.tex
\pdfoutput=1

\documentclass[11pt]{article}

\usepackage{acl}

\usepackage{times}
\usepackage{latexsym}

\usepackage[T1]{fontenc}

\usepackage[utf8]{inputenc}

\usepackage{microtype}

\usepackage{inconsolata}

\input{config}

\newcommand{\benchmark}{LangSuit$\cdot$E\xspace}
\newcommand{\prompt}[1]{{\fontfamily{lmtt}\selectfont #1}}
\newcommand{\model}[1]{{\fontfamily{qcr}\selectfont #1}}

\newcommand{\cmark}{\ding{51}}%
\newcommand{\xmark}{\ding{55}}%
\newcommand{\one}{\ding{202}\xspace}
\newcommand{\two}{\ding{203}\xspace}

\Crefname{table}{Tab.}{Tabs.}

\title{LangSuit$\cdot$E: Planning, Controlling and Interacting with Large Language Models in Embodied Text Environments}

\author{
Zixia Jia$^{1,}$\thanks{Equal contribution}, Mengmeng Wang$^{1,*}$,  Baichen Tong$^{1,}$, Song-Chun Zhu$^{1,2,3}$, Zilong Zheng$^{1,}$\thanks{Correspondence to Zilong Zheng \texttt{<zlzheng@bigai.ai>}} \\
$^1$ National Key Laboratory of General Artificial Intelligence, BIGAI \\
$^2$ Peking University
\quad $^3$ Tsinghua University \\ 
\url{https://github.com/bigai-nlco/langsuite.git}
}

\begin{document}


\maketitle

\begin{abstract}
Recent advances in Large Language Models (LLMs) have shown inspiring achievements in constructing autonomous agents that rely on language descriptions as inputs. However, it remains unclear how well LLMs can function as few-shot or zero-shot embodied agents in dynamic interactive environments. To address this gap, we introduce \benchmark, a versatile and simulation-free testbed featuring 6 representative embodied tasks in textual embodied worlds. Compared with previous LLM-based testbeds, \benchmark (i) offers adaptability to diverse environments without multiple simulation engines, (ii) evaluates agents' capacity to develop ``internalized world knowledge'' with embodied observations,  and (iii) allows easy customization of communication and action strategies. To address the embodiment challenge, we devise a novel chain-of-thought (CoT) schema, \model{EmMem}, which summarizes embodied states \wrt history information.
Comprehensive benchmark results illustrate challenges and insights of embodied planning.
\benchmark represents a significant step toward building embodied generalists in the context of language models.
\end{abstract}

\begin{figure*}[t!]
        \includegraphics[width=\textwidth]{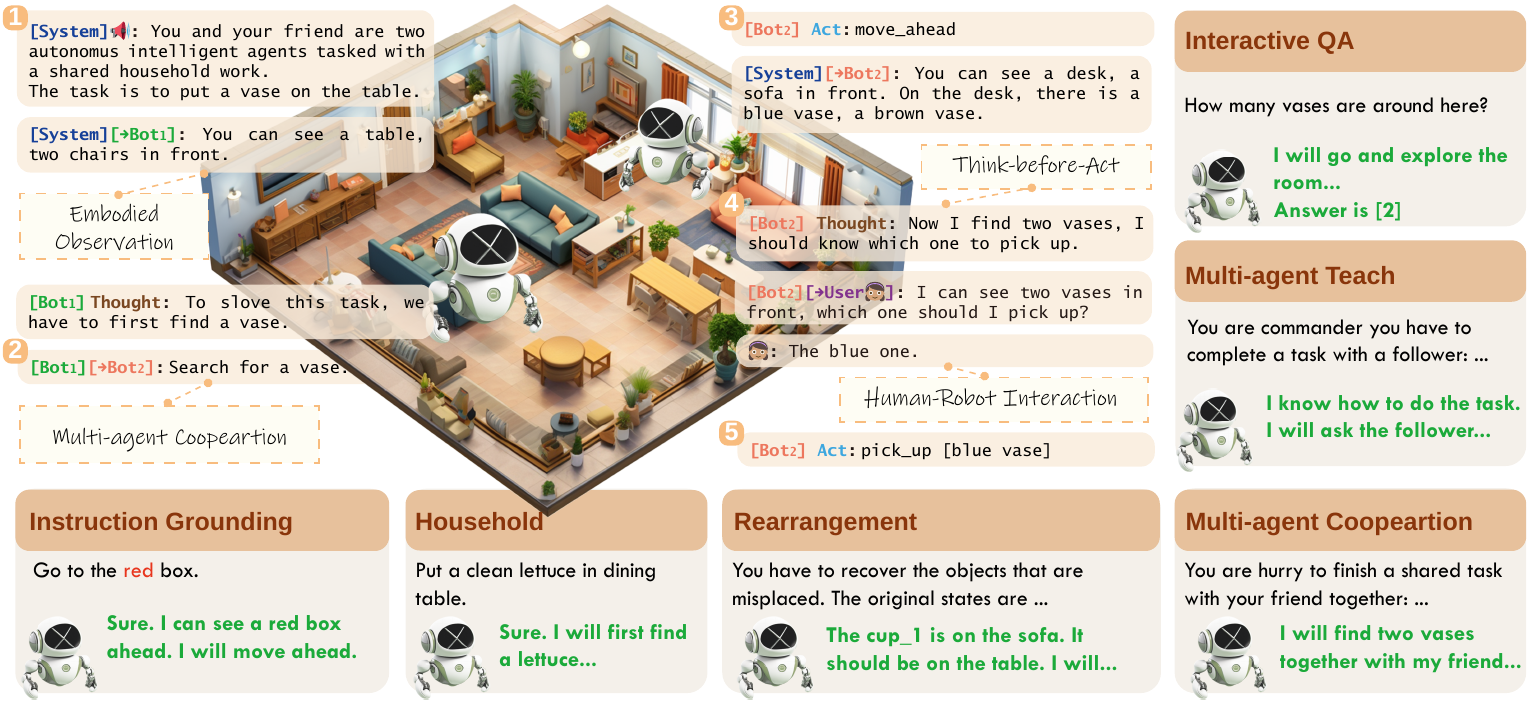}
    \caption{\textbf{Task illustration of \benchmark.} Top-Left: a typical example of a cooperative task between two agents. The agents are entirely blind (\ie, without visual perception) and can only obtain embodied information from the system. Around: Exemplar tasks supported by \benchmark.}
    \label{fig:teaser}

\end{figure*}

\section{Introduction}

Creating embodied agents that can perceive, think, and interact with simulated environments from the agents' own perspective is a long-standing yet imperative challenge within the realms of robots and artificial intelligence.
Recent striking progress of \acp{llm} has ushered promising avenues for constructing autonomous agents, wherein \acp{llm} serve as their central planners or controllers, evidenced by several intriguing demos and benchmarks within simulated environments~\citep{reed2022generalist, park2023generative,babyai_iclr19} or even realistic physical worlds~\citep{driess2023palm,huang2022inner} across a diverse spectrum of domains, spanning from web browsing~\citep{zhou2023webarena,yao2022webshop}, language games~\citep{park2023generative} and householding~\citep{ALFWorld20}, \etc. These language agents, reliant solely upon language descriptions as their inputs, can benefit from LLMs' exhibited commonsense knowledge, emergent semantic reasoning and problem-solving skills. Despite the promising advances of prior LLM-as-agent tasks, it remains unclear that:

\textit{Without perception error caused by visual models, can \acp{llm} behave as \textbf{embodied agents} to solve complicated tasks within dynamic environments?}

Previous works mostly treat \ac{llm} agents as task planners, providing them with the global observations and all required information (\eg, entire history) in natural language to produce the next step, which contrasts the nature of chat-based language models that can potentially form their internal map and history across conversational turns. Moreover, the action spaces across tasks are specially designed and can only be evaluated within specific environments, which limits the potential capacity of evaluating the generalizability across tasks. In stark contrast, research on embodied agents and environments has gone through many successful attempts -- numerous open-source embodied systems developed with realistic simulation engines, such as AI2Thor~\citep{ai2thor}, 
VirtualHome~\citep{puig2018virtualhome}, spurring the trend of learning embodied instruction-following challenges, such as navigation~\citep{chattopadhyay2021robustnav}, manipulations~\citep{ehsani2021manipulathor}, question answering~\citep{gordon2018iqa},\etc. Models learned within these environments, nevertheless, struggle with their dependency on simulation engines and are therefore limited to specific tasks.
Other prominent research works such as PaLM-E~\citep{driess2023palm} and RT~\citep{zitkovich2023rt} sought to powerful vision-language models~(VLMs) for embodied scene perception and language models for low-level control. However, these works can only be applied to constrained environments and require large volumes of data for pre-training. 

To fill the gap and step towards general embodied intelligence, we introduce \textbf{\benchmark}, a systematic and simulation-free testbed for evaluating \acp{llm}' capabilities across different embodied tasks in textual embodied worlds. As listed in \Cref{tab:langsuite_tasks}, \benchmark showcases 6 representative embodied tasks, with variants of the number of rooms, the number of agents, the action spaces of agents (whether they can communicate with each other or ask humans). Compared with prior LLM-based agent testbed, \benchmark presents superior benefits on building and benchmarking \acp{llm}' embodied capabilities. \textbf{First}, due to the nature of textual worlds, \benchmark incorporates a general yet fully customizable indoor environment that can easily adapted to most simulated embodied environments, (\eg, AI2Thor~\citep{ai2thor}, VirtualHome~\citep{puig2018virtualhome}), without resorting to variants of simulation engines and benchmark versions. \textbf{Second}, \benchmark only presents embodied observations to agents at each step. A typical example is shown in \Cref{fig:teaser}, where only in-front observations within certain distances (unblocked by walls) are presented to agents at each step. Prior study~\citep{huang2022inner} has shown that language models can potentially form ``internalized world knowledge'' through thinking in language. For instance, they shall potentially know that executing ``\prompt{turn\_right by 90 degrees}'' four times shall remain in the same direction as the beginning. Such capacity could potentially enable \ac{llm} agents to reason over action histories and observations to plan for the next. \textbf{Third}, the agents in \benchmark are fully-customizable \wrt their action spaces and communicative capabilities, \ie, one can easily adapt the communication and acting strategy from one task to another. 

\begin{table*}[t!]
    \centering
    \caption{\textbf{Statistics of Tasks Integrated in \benchmark.} We provide a unified and simulation-free environment that can easily adapt to most embodied tasks and benchmarks. }
    \label{tab:langsuite_tasks}
    
    \resizebox{\linewidth}{!}{%
    \begin{tabular}{llcccccc}
    \toprule
       \textbf{Task Type}  &  \textbf{Simulator/Env.} & \textbf{\# of Scenes} & \textbf{\# of Tasks} & \textbf{\# of Actions} & \textbf{Multi-Room} & \textbf{Multi-Agent} & \textbf{Communicative} \\ \midrule
          Instruction Grounding (IG) & Mini Grid/BabyAI~(\citeyear{babyai_iclr19}) & 105 & 500 & 6 & \cmark & \xmark & \xmark \\
       Rearrangement & AI2THOR/Rearrangement~(\citeyear{luca2021rearrangement}) & 120 & 500 & 8 & \xmark & \xmark & \xmark  \\
         Interactive QA (IQA) &  AI2THOR/IQA~(\citeyear{gordon2018iqa}) & 30 & 3000 & 5 & \xmark & \xmark & \cmark  \\
         Household &  AI2THOR/ALFred~(\citeyear{ALFRED20}) & 120 & 506 & 12 &  \xmark & \xmark & \xmark\\
         Multi-agent Household (MA-Household)& AI2THOR/TEACh~(\citeyear{teach2022}) & 120 & 200 & 13 & \xmark & \cmark & \cmark \\
         Multi-agent Watch-and-Help (MA-WAH)& VirtualHome/Co-WaH~(\citeyear{zhang2023building})& 2 & 50 & 6 & \cmark & \cmark & \cmark \\
    \bottomrule
    \end{tabular}    
    }

\end{table*}

To evaluate the current prevalent LLMs's performance on \benchmark, we derive a comprehensive benchmark across 6 different tasks. Of note, annotating or generating golden trajectories is commonly labor-intensive and time-consuming. In \benchmark, we can easily form a benchmark by leveraging and modifying existing annotations, a by-product benefit of pursuing such a general textual embodied world. An underlying hypothesized benefit is that knowledge learned from one task can also enhance the performance of the others. 

Lastly, inspired by grid cells in neural science~\cite{banino2018vector} and the ``lost-in-middle'' observation of long-context language models~\cite{liu2023lost}, we propose \model{EmMem}, an \textbf{Em}bodied planning strategy with an emergent embodied \textbf{Mem}ory. Specifically, the language model learns to summarize and predict their embodied states in an in-context manner given long-context trajectories and observations. Results on the Household task present the effectiveness of the \model{EmMem} strategy.



In summary, this paper makes the following contributions: i) we deliver a systematic and useful testbed with 6 exemplar tasks for testing and training LLMs as embodied agents; ii) We introduce \model{EmMem}, a novel prompting strategy for embodied tasks; iii) We show substantive benchmark results and call for future exploration in \benchmark.

\section{Related Work}

\begin{figure}
    \centering
    \resizebox{\linewidth}{!}{
    \includegraphics{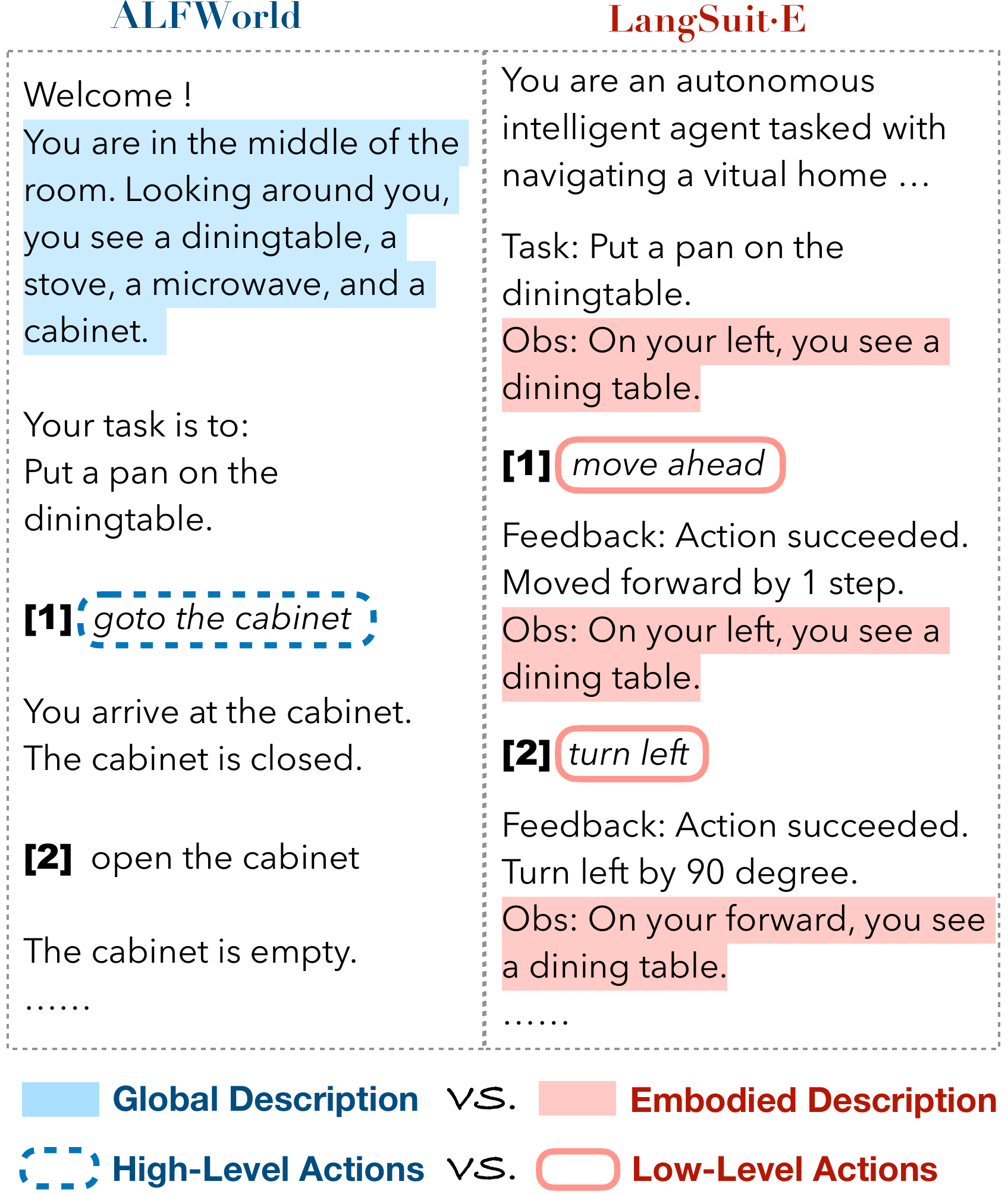}}
    \caption{\textbf{Comparison of our Embodied Environment Description with ALFWorld~\cite{ALFWorld20}.}}
    \label{fig:comp_alfworld}
\end{figure}

\paragraph{Evaluating LLM as Autonomous Agents}
The concept of LLM-as-Agent has triggered the development of many systematic and interactive environments that allow for textual interactions and show well-aligned language groundings in visual or physical modalities. 
TextWorld~\citep{cote18textworld} is one of the earliest testbeds that presented a text-based game sandbox for training and testing \ac{rl} agents. Based on this framework, ALFWorld~\citep{ALFWorld20} proposed an embodied challenge that aligns the ALFRED~\cite{ALFRED20} benchmark to TextWorld. However, this environment still requires a simulator to generate text descriptions and simplifies the embodied challenge by presenting full world observations to agents (\cref{fig:comp_alfworld}).
More recently, enhanced by harnessing the capacities of chat-based \acp{llm}, a variety of language agents targeting a spectrum of different tasks have been demonstrated. For example, WebShop~\cite{yao2022webshop}, Mind2Web~\cite{deng2023mind2web} and WebArena~\citep{zhou2023webarena} present environments on open-domain web shopping and browsing;
Agents~\citep{zhou2023agents} design a library that supports long-short term memory; AgentBench~\citep{liu2023agentbench} presents a comprehensive evaluation of 25 different \acp{llm} across 8 environments. 
In addition to single-agent settings, recent advances also show emergent behaviors in multi-agent \ac{llm} systems~\citep{gong2023mindagent,park2023generative,chen2023agentverse,wang2023avalon}. However, most of these tasks over-simplify the agent's observations and actions by presenting with full knowledge of the world's status.

\paragraph{LLMs for Embodied Tasks and Robotics}


The rise of \acp{llm} has spurred many inspirations on conventional embodied and robotics systems. 
\cite{vemprala2023chatgpt} presented an empirical success of using ChatGPT in zero-shot robotics controls. 
SayCan~\citep{ahn2022can} and LLM-Planner~\citep{song2022llm} employ visual navigation to collect information in the house for the challenging grounded plan generation.
Another thread of works presents attempts on multimodal embodied environments by leveraging a visual-language pretraining model for world perception and \acp{llm} for planning. For instance, many works leverage \acp{llm} to enhance agent ability in the Minecraft environment~ \citep{wang2023voyager,zhu2023ghost,reed2022generalist, baker2022video,fan2022minedojo}. 
\citet{wu2023embodied} construct a multimodal dataset containing triplets of indoor scenes, instructions, and action plans, where they employ the AI2-THOR simulator~\citep{kolve2017ai2} to expand indoor scenes as environments and leverage GPT-3.5 to automatically annotating action plans. However, due to the imperfect multimodal understanding and dynamic scene environments, these tasks are limited to single-agent settings and simple manipulations and navigation in domain-specific environments. Differently, \benchmark presents a simulation-free and embodied framework that allows a variety of environments and agent settings.

\paragraph{Planning and Acting with LLMs}

Inspired by the success of \ac{cot} on language reasoning tasks, many works propose to enhance the planning and acting capabilities of \acp{llm} in a similar manner. ReAct~\citep{yao2022react} proposes to combine \ac{cot} reasoning and actions to enhance the efficiency of language models.
\citet{dasgupta2023collaborating} investigate a Planner-Actor-Reporter agent paradigm, where the Planner is equipped with a language model to perform reasoning. 
\citet{wu2023plan} propose a Plan, Eliminate and Track framework, aiming to mitigate constraints, such as fine-tuning inefficiency and bias from pre-training, of LLM with transformer architecture. Related to us, \citet{pan2023langnav} transfers visual perception from the virtual environment to language descriptions, then prompts LLM to solve the specific navigation tasks only leveraging language descriptions. But different from them, we provide a unified embodied environment that is capable of real-time feedback and integrates various embodied tasks.

\section{The \benchmark Testbed}

The goal of \benchmark is to provide a unified testbed thereby evaluating the diversified ability of LLM-as-Agents in a dynamic embodied environment with textualized perceptual representation and realistic physical limitation. Moreover, by unifying various embodied tasks in one environment, agents could improve the ability of knowledge transfer and construct consistent world cognition from interactions with the environment to solve different kinds of tasks.

\subsection{Generic Embodied Task Planning} 
\benchmark aims to construct a general textual embodied world by defining a lightweight and simulation-free environment system. 
The embodied task of \benchmark can be formulated as a partially observable Markov decision process~(POMDP) $(\mathcal{S}, \mathcal{A}, \mathcal{T}, \mathcal{R}, \mathcal{U}, \mathcal{O})$, which encompasses embodied agent state space $\mathcal{S}$, action space $\mathcal{A}$, deterministic transition function $\mathcal{T}: \mathcal{S} \times \mathcal{A} \rightarrow \mathcal{S}$, reward function $\mathcal{R}$, task instruction space $\mathcal{U}$ and embodied observation space $\mathcal{O}$.

\begin{table}[t!]
\small
\centering
\caption{\textbf{Examples of different action settings in \benchmark.} Refer to Appendix for the complete set.}

\resizebox{\linewidth}{!}{%
\begin{tabular}{l|p{5cm}}
\toprule
\textbf{Action Type} & \textbf{Description} \\ \midrule
\multicolumn{2}{l}{\textit{Low-level actions}} \\ \midrule
\prompt{move\_ahead} & Move ahead by 1 step. \\
\prompt{turn\_left} & Turn left by 90 degrees. \\
\prompt{turn\_right} & Turn right by 90 degrees. \\
\prompt{pick\_up [obj]} & Pick up an object identified by \prompt{[obj]}. \\
\prompt{put [obj, recep]} & Put the object \prompt{[obj]} into the receptacle \prompt{[recep]}. \\ 
\prompt{open [obj]} & Open object \prompt{[obj]} if it is openable. \\
\prompt{close [obj]} & close object \prompt{[obj]} if it is closeable. \\
\prompt{stop [answer]} & Stop making actions and conclude the final answer in \prompt{[answer]}.\\ 
... & ...\\
\midrule
\multicolumn{2}{l}{\textit{High-level actions}} \\ \midrule
\prompt{goto [loc]} & Go to a location that is in current observation. \\ 
... & ...\\

\midrule
\multicolumn{2}{l}{\textit{Communicative actions}} \\ \midrule
\prompt{chat [message]} & Send messages to other agents. \\
\prompt{ask [message]} & Send queries to users.\\
\bottomrule
\end{tabular}
}
\label{tab:action_type}
\end{table}

\paragraph{Embodied States} 
The state refers to the representation of the environment that an agent is in at a given time. 
\benchmark encompasses diverse indoor scene layouts. The state of our embodied setting includes the indoor arrangement, object information, the history of the previous action decisions and executive outcomes, the overall target or programmatic sub-goals, the limitations that must be followed, and so on.

\paragraph{Embodied Observations}
To simulate real-world conditions, the environment is partially observable, which means that an agent is not capable of perceiving the entire world or environment states. Based on this, \benchmark provides the embodied observations of partial objects in the environment with full customization of each agent's view range.
In particular, \benchmark includes a module dedicated to describing environmental objects that are perceptible within an agent's uniquely configured visual scope using natural language. For instance, the language expressions could be ``\prompt{You can see a red ball on your right}'' or ``\prompt{You can see nothing ahead}''.

\paragraph{Embodied Actions}
\benchmark incorporates a wide spectrum of executable embodied actions, covering most actions typically encountered within various embodied environments. The agent decides to execute a specific action in each time step according to the embodied state and its internal world knowledge.
Tab. \ref{tab:action_type} presents an enumeration of the action types that are currently accommodated in the \benchmark. Refer to Tab. \ref{tab: complete_action_type} in Appendix for the complete action set.
Furthermore, \benchmark exhibits substantial flexibility by allowing for the customized configuration of action scopes. For example, the movement distance associated with the \texttt{move\_ahead} action can be readily modified by adjusting the relevant hyperparameters, thereby facilitating tailored adjustments to suit specific embodied task requirements.

\paragraph{Feedback} \benchmark environment conveys feedback through the use of natural language descriptions.
This feedback is automatically generated by our system by evaluating the feasibility of the action predicted by LLMs in the context of the current state. The evaluation takes into account physical constraints imposed by both the agent and the environment, such as the agent's grasping range or environmental obstacles, ensuring that the action aligns with the agent's physical capabilities and the environmental context; refer to \cref{fig:langal} and Appendix for examples.



\subsection{Communications in \benchmark}

The platform additionally provides multi-agent and human-agent communication channels to enable multi-agent cooperation and human-in-the-loop tasks. The \textit{Communicative actions} in \cref{tab:action_type} are additional actions to be appended for this capacity.


\vspace{0.1cm}
\paragraph{(Multi-)Agent Interactions}
Multi-agent cooperations are widely explored in many recent LLM-as-agent tasks~\cite{zhou2023agents,park2023generative}. In \benchmark, the agents are fully customizable regarding their action spaces and communicative capabilities, \ie, one can easily adapt the communication and acting strategy from one task to another. See \cref{fig:teaser} for a typical example.

\vspace{0.1cm}
\paragraph{Human Communication}
\benchmark supports human-agent interactions through two primary mechanisms: firstly, it empowers humans to control an agent or play the role of the agent to cooperate with other agents. Secondly, it affords agents the capability to seek clarification from humans in instances of environmental or instructional ambiguity, enabling humans to offer assistance by imparting pertinent information via a communication protocol, thus enhancing task completion.

\subsection{Benchmark \benchmark Tasks}

We integrate a diverse range of embodied tasks within the unified \benchmark testbed. \cref{tab:langsuite_tasks} demonstrates the specific categories of tasks that have been accommodated within \benchmark. We carefully select the six tasks to illustrate how our testbed can be applied to single-agent embodied understanding and planning (Instruction Grounding, Interactive QA, Household and Rearrangement), multi-agent cooperation (MA-WAH), multi-agent teaching (MA-Household), each of which represents a significant challenge within the current embodied community. 
We re-implement all tasks in our embodied environments to reach the objective of a unified simulation-free testbed.
Notably,
\benchmark has an inherent capacity for facile expansion and adaptation to accommodate additional embodied tasks with minimal resource consumption, for the reason of the wealth meta-capabilities it possesses, encompassing easy customization of communication and action strategies and the ability to support interactions involving multiple agents, among others.

\vspace{0.1cm}
\paragraph{Task Instructions} 
Task instruction provided to an agent focuses on the characteristics and constraints of tasks with a particular type, involving customized descriptions of the specific type, action spaces of tasks, and optional few-shot examples to show the diverse execution processes of some tasks under the same type.

\benchmark provides a template for the automated generation of diverse task instructions. By filling out customized task descriptions, definitions of action spaces pertinent to a particular task type, or the incorporation of few-shot illustrative examples, \benchmark is capable of generating distinct sets of instructions that serve as prompts for LLMs regarding different tasks. The tasks that the current version of \benchmark supports:

\underline{\textit{Task I. Instruction Grounding (IG)}}~\cite{babyai_iclr19} provides synthetic
natural-looking instructions (\eg, ``\prompt{put the red ball next to the box on your left}''), requiring the agent
to navigate the world (including unlocking doors) and move objects to specified locations.

\underline{\textit{Task II. Rearrangement}}~\cite{luca2021rearrangement} involves moving and modifying (\ie, opening/closing) randomly placed objects within a room to obtain a goal configuration.

\underline{\textit{Task III. Interactive QA (IQA)}}~\cite{gordon2018iqa} is the task of answering questions that require the agent to interact with a dynamic environment, which contains three different question types:  Existence questions (``\prompt{Is there an apple in the kitchen?}''), Counting questions (``\prompt{How many forks are present in the scene?}''), and Spatial Relationship, called Contains questions (``\prompt{Is there lettuce in the fridge?}'').

\underline{\textit{Task IV. Household}}~\cite{ALFRED20} tasks contain both high-level goals like ``\prompt{Rinse off a mug and place it in the coffee maker.}'' and low-level language instructions like ``\prompt{Walk to the coffee maker on the right.}''

\underline{\textit{Task V. Multi-agent Teach (MA-Teach)}}~\cite{teach2022} tasks should be completed with the cooperation of two agents. A Commander with access to oracle information about a task communicates in natural language with a Follower. The Follower navigates through and interacts with the environment to complete tasks varying in complexity from ``\prompt{make coffee}'' to ``\prompt{prepare breakfast}'', asking questions, and getting additional information from the Commander.

\underline{\textit{Task VI. Multi-agent Coopeartion}} tasks focus on evaluating the cooperation and communication ability of multiple agents to accomplish shared goals. We follow C-WAH~\cite{zhang2023building} to use a watch-and-help~(WAH) configuration in this task. Specifically, all agents shall negotiate with each other to break down the task based on their individual observations and capabilities.
Different from C-WAH, all agents in this task can only perceive their own embodied observations.


\vspace{0.1cm}
\paragraph{Gold Trajectories Generation} 
We have implemented an extra expert module for task I-IV~(the multi-agent settings require agent interactions), dedicated to generating gold trajectories and affording us the capability to construct training datasets that may be subsequently employed for the potential fine-tuning of LLMs.
Given a task without gold trajectories, our expert module automatically produces a reasonable and executable trajectory, including every-step actions. Specifically, we first analyze a list of objects necessitating manipulation according to the answer or the final state of the task. Second, we construct a list of sub-tasks corresponding to each operation of an object and carefully devise the execution sequence of sub-tasks. Finally, we adopt the A-star algorithm to chart the shortest paths between the dynamic agent location (the agent's location after finishing the last sub-task) and the target object for each sub-task. As a consequence, our expert module outputs an exhaustive sequence of actions meticulously crafted to fulfill the task at hand.



\subsection{Evaluation Metrics} 
\label{sec:eval}
We follow metric design schemas as in \cite{luca2021rearrangement,ALFRED20} to provide the following evaluation metrics:
\begin{itemize}[leftmargin=*, noitemsep, topsep=5pt]
    \item \textbf{Success Rate (SR)}: the ratio of tasks in which an agent achieves the target goal or state.   
    \item \textbf{Goal-Conditioned Success Rate (Goal-SR)}: the sub-goal completion ratio of a task, averaged across all the tasks of the same type.
    \item \textbf{Misplaced Rate (Misplaced)}: the misplaced objects in the \textit{Rearrangement} task, which computes the ratio of misplaced objects at the end of the task, denoting better performance with a lower value.
    \item \textbf{Fixed Strict Rate (Fixed Strict)}:  the correctly reset objects in the \textit{Rearrangement} task, which computes the ratio of correctly reset objects at the end.
    \item \textbf{Accuracy (Acc)}: the proportion of correct answers compared to the total number of questions. The metric solely evaluates on the \textit{IQA} task. 
    \item \textbf{Average Step (Avg Steps)}: the average number of actions an agent executes to complete all the tasks.
    \end{itemize}
















\section{\model{EmMem}: Interactive Planning with \underline{\model{Em}}bodied \underline{\model{Mem}}ory}

\begin{figure*}[t!]
    \centering
    \small
    \resizebox{\linewidth}{!}{
    \includegraphics{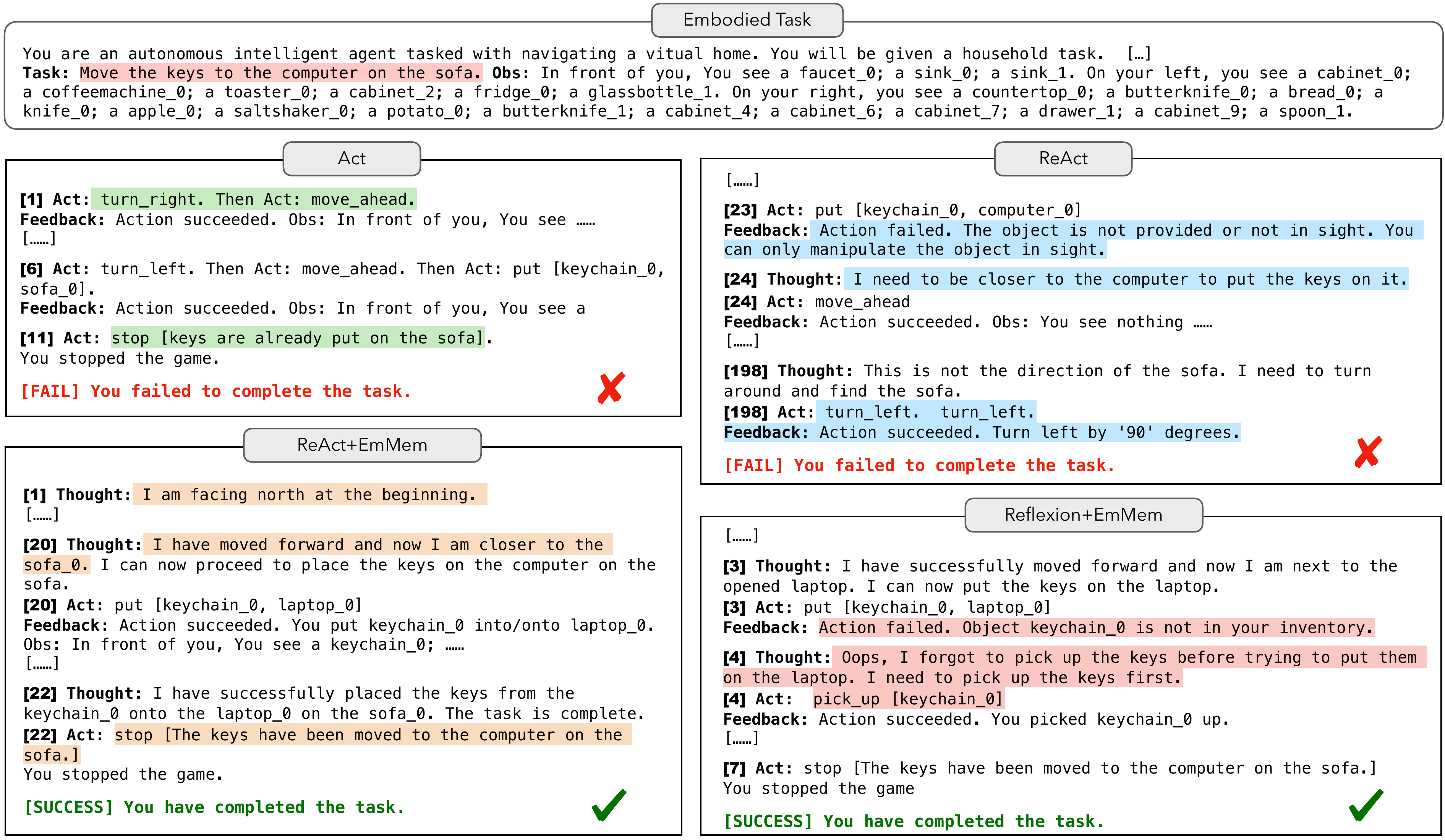}}
    \caption{A case study for four prompt strategies on Household. \model{Act}: The model predicts the next action only. \model{ReAct+EmMem}: Our EmMem strategy with \model{ReAct} strategy (Reason+Act) \cite{yao2022react}.  \model{Reflexion+EmMem}: Our EmMem strategy with \model{Reflexion} strategy \cite{shinn2023reflexion} (1 additional trail of verbal reflection). Because \model{Reflexion} will try an additional trail if the task fails and the summarization of the whole last trail will be input to the new trail planning through the Task prompt, we do not show the case in this figure. More details and 
    the prompt instructions for these strategies can be found in Appendix \ref{app:prompt}.}
    \label{fig:langal}
\end{figure*}

One \textit{de facto} paradigm of prompting LLMs for planning tasks is to leverage chain-of-thought~(CoT)~\cite{wei2022chain} strategy which carefully designs and selects cases of reasoning traces as in-context examples. \model{ReAct}~(Reason+Act)~\cite{yao2022react}, one prior success in a couple of reasoning and planning tasks, generates both \textit{reasoning traces} and actions through an extra \textbf{Thought} process. Following this, \model{Reflexion}~\cite{shinn2023reflexion} proposes to verbally reflect on task feedback signals, then maintain reflective text in a memory buffer in subsequent trials. However, all these strategies failed to reveal satisfactory results in our embodied environments; refer to \cref{fig:langal} for typical failures.

We present \model{EmMem}, a novel embodied planning paradigm that reasons and summarizes the agent's current embodied memory in an in-context and emergent manner. Specifically, the LLM agent has to explicitly predict and summarize their current embodied state before proceeding to the actual planning and action prediction step. We provide two underlying rationales for such prompts: \textbf{First}, as for mammals' embodied navigation, neurons like grid cells integrate path information and are thought to be critical in self-motion~\cite{banino2018vector,hafting2005microstructure}. The \model{EmMem}, auto-regressively in principle, produces integrated embodied information as grid cells. \textbf{Second}, recent literature has observed the lost-in-middle performance for long-context language models~\cite{liu2023lost}. Our \model{EmMem} enhances the ``middle'' information in a residual-like approach~\cite{he2016deep}. \cref{fig:langal} illustrates a case study of \model{EmMem} and other strategies.


        


\begin{table*}[t!]
    \centering
    \small
    \caption{Evaluation results on the \textit{test} set of single-agent  
 task types in \benchmark. The evaluation results on Household task in terms of Path Weighted Metrics are in Appendix \ref{app:path}. }
    \label{tab:results_of_100_items}
    \resizebox{\textwidth}{!}{%
    \begin{tabular}{l|cc|cccc|cccc|ccc}
    \toprule
        \multirow{2}{*}{\textbf{Model}} &  \multicolumn{2}{c|}{\textbf{IG}} & \multicolumn{4}{c|}{\textbf{Rearrangement}} & \multicolumn{4}{c|}{\textbf{IQA}} & \multicolumn{3}{c}{\textbf{Household}}  \\

        & \textbf{SR.} & \makecell{
    \textbf{Avg} \\ \textbf{Steps.}↓} & \textbf{SR.} &  \textbf{Misplaced.}↓ &  \makecell{\textbf{Fixed} \\ \textbf{Strict.}} & \makecell{\textbf{Avg} \\ \textbf{Steps.}↓} & \makecell{\textbf{Exists} \\ \textbf{Acc.}} & \makecell{\textbf{Contains} \\ \textbf{Acc.}} & \makecell{\textbf{Counts}  \\ \textbf{Acc.}} & \makecell{\textbf{Avg} \\ \textbf{Steps.}↓} & \textbf{SR.} &  \textbf{Goal-SR.} & \makecell{\textbf{Avg} \\ \textbf{Steps.}↓}  \\
        
        \midrule
       \textit{Low-Level Actions Setting}
       \\\midrule
        \prompt{Act}  & 13.0 & 28.2 & 8.0 & 115.3 & 17.6 & 12.4 & 14.9 & 36.0 & 10.9 & 16.7 & 9.0 & 21.5  &  22.8\\
        \prompt{ReAct}  & 12.0 & 25.3& 5.0 & 117.9 & 12.3 & 8.4 & 8.9 & 18.3 & 2.1 & 13.2 & 14.0 & 26.7 & 44.7 \\
        \prompt{ReAct-EmMem}  & 11.0 & 22.8 & 6.0 & 108.3 & 13.7 & 9.5 & 48.0 & 58.0 & 6.0 & 14.8 & 10.0 & 26.3 & 28.5 \\
        \prompt{Reflexion}  & 20.0 & 29.9 & 8.0 & 99.3 & 14.3 & 13.7& 62.0 & 72.0 & 38.0 & 23.7 & 19.0 & 34.6 & 28.6\\
        \prompt{Reflexion-EmMem}  & 16.0 & 36.8 & 10.0 & 100.4 & 15.7 &12.5 & 60.0 & 86.0 & 72.0 & 20.7 & 14.0 & 32.0  & 35.1 \\
        \midrule
       \textit{High-Level Actions Setting}
       \\\midrule
        \prompt{Act} & 77.0 & 6.9 & 5.0 & 112.3 & 17.2 & 7.0 & 24.0 & 76.0 & 30.0 & 10.3 &  23.0 & 37.8 &  16.2 \\
        \prompt{ReAct} & 84.0 & 14.3 & 7.0 & 111.2 & 15.2 & 6.9 & 44.0 & 88.0 & 18.0 & 15.1 & 22.0 & 42.0  & 21.4 \\
        \prompt{ReAct-EmMem} & 85.0 & 16.9 & 6.0 & 109.7 & 11.9  & 9.0 & 32.0 & 80.0 & 14.0 & 13.1 & 32.0 & 47.8  & 35.2 \\
        \prompt{Reflexion}  & 86.0 & 14.4 & 8.0  & 102.7 & 17.5  &  6.5 & 44.0 & 90.0 & 22.0 & 9.9 & 33.0 & 53.0  & 27.7 \\
        \prompt{Reflexion-EmMem}  & 86.0 & 15.7 & 8.0 & 100.22 & 16.2 & 8.9 & 42.0 &  80.0 & 20.0 & 16.6& 39.7 & 53.3 & 34.1\\
    \bottomrule
    \end{tabular}
    }

\end{table*}

\begin{table}[t!]
\caption{Evaluation results on the \textit{test} set of multi-agent  
 task types in \benchmark.}
    \resizebox{\linewidth}{!}{%
    \begin{tabular}{l|cc|cc}
    \toprule
         \multirow{2}{*}{\textbf{Model}} &\multicolumn{2}{c}{\textbf{MA-Household}}  & \multicolumn{2}{c}{\textbf{MA-WAH}}  \\
         & \textbf{SR.} &  \textbf{Goal-SR.} & \textbf{SR.} &  \textbf{Goal-SR.} \\
       \midrule
        \textit{High-Level Actions Setting}
       \\\midrule
        \prompt{Act} & 18.00 & 44.80 & 36.00 & 70.17  \\
        \prompt{ReAct} & 18.00 & 60.63 & 38.00 & 78.20 \\
        \prompt{ReAct-EmMem} & 18.00 & 64.74 &  46.00     & 76.13   \\
        \prompt{Reflexion}  & 18.00    & 66.58 &    56.00   &  88.43  \\
        \prompt{Reflexion-EmMem}  & 20.00    & 65.81 &  64.00  &  90.37 \\ 
    \bottomrule
    \end{tabular}
    }
\label{tab:multi}
\end{table}

\section{Experiments}

\subsection{Setup}
We evaluate the ability of popular LLMs as agents to solve embodied tasks in our embodied textual environment. It is crucial to note that our testbed does not provide any global information, necessitating that agents mobilize more capabilities in the more realistic and complicated embodied environment to execute tasks, including comprehension of embodied description, retention of historical exploration within the environment, and self-awareness of its current state. 
We mainly use\footnote{We did not fully test some open-source LLMs with small sizes, such as LLaMA, because we made a preliminary attempt to utilize LLaMA 2-7B to solve Instruction Grounding and IQA tasks but found it failed in all tasks} \prompt{gpt-3.5-turbo} (with temperature being 1) to evaluate on \benchmark. \textbf{Notably}, for 
API-cost economy, 
We evaluated 100 randomly sampled tasks for IG, Household, and Rearrangement, along with 50 randomly sampled tasks for each sub-task of IQA (Exists, Contains, and Counts), MA-Household, and MA-WAH. We conduct experiments both in the low-level (fine-grained actions) and high-level action settings. The evaluation metrics adopted mainly are introduced in \cref{sec:eval}. 

\subsection{Main Results and Observations}
In this section, we summarize the empirical results and focus on addressing the following questions:

\vspace{0.1cm}
\paragraph{Q1: How do previous CoT strategies perform in \benchmark?} 

\cref{fig:langal} and \cref{tab:results_of_100_items} show the qualitative and numerical results of previous CoT methods on single-agent tasks. Our observations can be summarized as follows: \one~\model{Act}-only \textit{vs.} \model{ReAct} \textit{vs.} \model{Reflexion}:  The \model{Act} can not strictly follow our desired format (it generates two actions in each step, where we only require the first one), and it predicts the \textit{stop} action even though the task is not completed. Differently, \model{ReAct} has the capability of considering the correct plans after the action failed, but explores more steps, resulting in exceeding our step limit (in the low-level action setting on Household). Notably, \model{ReAct} drops the performance seriously in the IQA tasks in low-level action setting. A possible reason is that CoT causes more severe hallucination problems, as illustrated in \citet{vu2023freshllms}. 
The \model{Reflexion}, which goes through more trials and self-reflection process, showcases significantly better performance on most task types. And equipped with EmMem, the performances are further improved in some cases.
\two~Low-level action \textit{vs.} high-level actions: We compare the high-level action sets with low-level actions introduced in \cref{tab:action_type}. In a high-level setting, an agent does not need to know how to perform low-level action sequences to achieve a goal place/object. Empirically and intuitively, we can observe that high-level actions make embodied tasks relatively easy to solve (fewer steps and higher SR.), which indicates the challenges of our fully embodied action setting. Although prompts with \textbf{Thought} (Reflexion, EmMem) increase the average execution steps of GPT-3.5, the performances of GPT-3.5 are improved in most cases. It is surprising that the \textbf{Thought} process drops the Fixed Strict scores in the low-level actions setting on Rearrangement. 

\vspace{0.1cm}
\paragraph{Q2: How does \model{EmMem} help in embodied planning tasks?} 
As shown in \cref{tab:results_of_100_items} and case results \cref{fig:langal}, with reasoning and summarizing the current embodied state in the \model{ReAct+EmMem} setting, an agent is more clear about the state of itself and the execution process of a task, hereby successfully carrying out the task; With both reflexion and memory (\model{Reflexion+EmMem}) strategies, an agent can recognize its mistakes quickly and reformulate following plans efficiently.
We also observe that \model{EmMem} is more useful under the low-level action settings compared with high-level ones. Such observation is in line with the intuition that low-level settings can benefit more from embodied memory. 

\begin{figure}[t!]
    \centering
    \includegraphics[height=5.6cm]{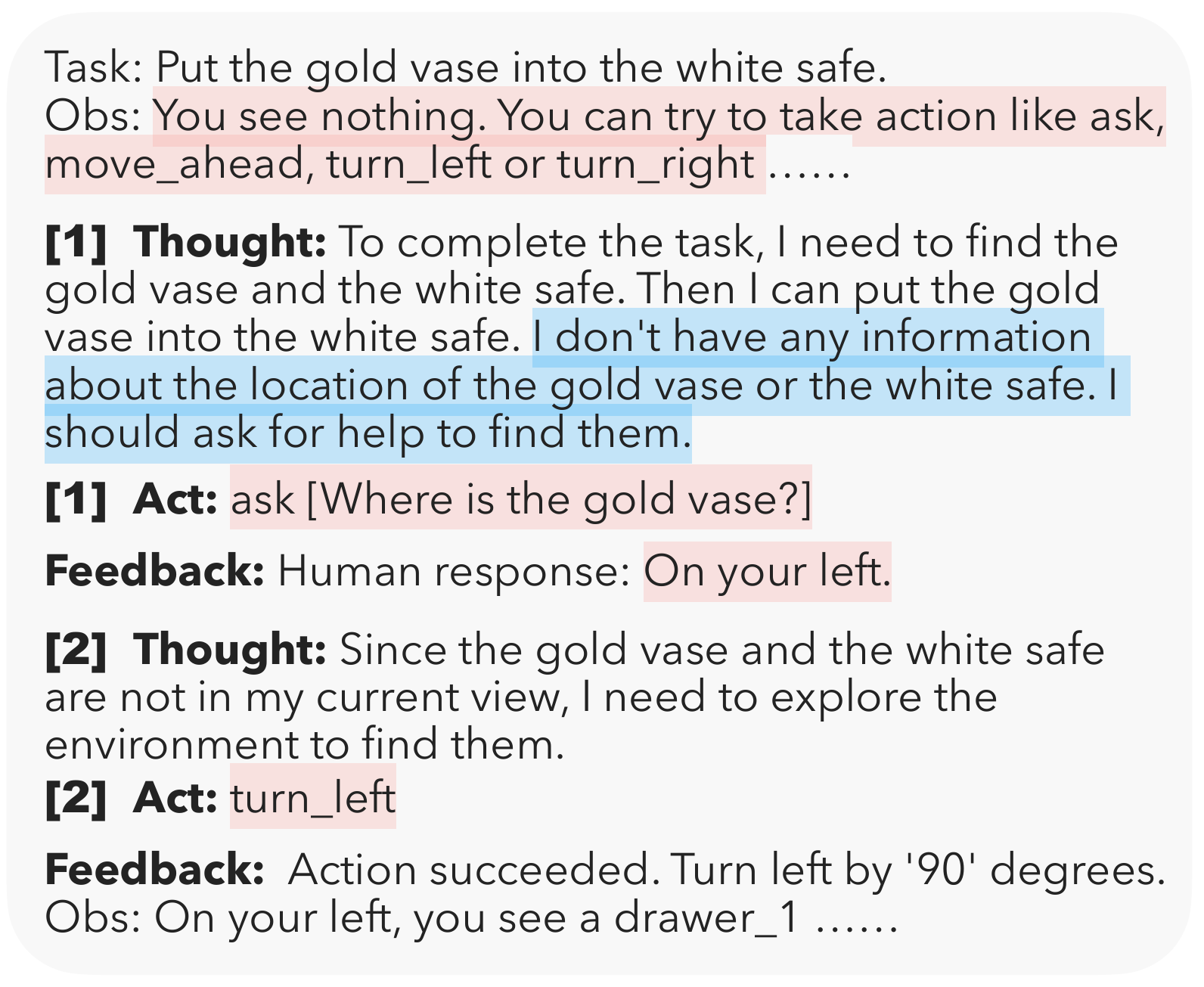}
    \caption{Demonstration of human communications.}
    \label{fig:human_chat}
\end{figure}

\vspace{0.1cm}
\paragraph{Q3: How can human-in-the-loop be helpful?} 
Previous works have shown that appropriate human-agent communication can substantially improve the efficiency in embodied grounding~\cite{huang2022inner}. In \benchmark, we also incorporate the ask-for-help setting to obtain human preferences and selection, especially when it comes to ambiguous task instructions (a qualitative demonstration in \cref{fig:human_chat}).
\vspace{0.1cm}
\paragraph{Q4: Does the temperature highly influence the EmMem performance?} 
We conducted experiments with different temperature settings for GPT-3.5 and reported the corresponding standard deviations in \cref{tab:temp}. We have observed consistent performance across various temperature settings. Our hypothesis is that the CoT strategy, coupled with exemplar cases, plays a crucial role in guiding LLMs to follow a consistent reasoning trajectory. This observation underscores the effectiveness of our approach in maintaining stability and reliability in the model's performance.

\paragraph{Q5: How does the \model{EmMem} perform in the Multi-agent settings?} 
We only use high-level actions in multi-agent tasks to reduce task complexities. We observe that all API-based agents maintain the capacity of multi-agent cooperations with relatively high Goal-SRs. We did not report low-level results as there is barely a successful case. As demonstrated in \cref{tab:multi}, \model{EmMem} achieves consistent improvement, except for the \textbf{Goal-SR} metric in MA-Household with Reflexion. Note that GPT-3.5 achieves high performance in MA-WAH tasks in most of the prompt strategies, because the feedback of each action provided in MA-WAN contains the self-state of agents. It is consistent with our intuition for designing \model{EmMem}: explicitly predicting and summarizing agents' current embodied state before proceeding to the actual planning and action prediction step is beneficial. It is exciting that even with the self-state of agents the environment provided, \model{EmMem} inspiring the self-awareness of agents' current stat still further improves the performance.

\begin{table}[t!]
    \centering
    \small
    \caption{Evaluation results of varying temperatures (\textbf{Temp}) on Household using Reflexion-EmMem strategy in high-level actions setting.}
    \label{tab:temp}

    \begin{tabular}{l|ccc}
    \toprule
       \textbf{Temp} & 0.5 & 1.0 & 1.5 \\
        
        \midrule
        \prompt{\textbf{SR.}}  & 40.00$\pm$3.46 & 39.67$\pm$0.58 & 40.67$\pm$0.58 \\
        \prompt{\textbf{Goal-SR.}}  & 52.47$\pm$3.71 & 53.30$\pm$2.43 & 52.86$\pm$1.13
\\
       
    \bottomrule
    \end{tabular}
\end{table}



        



\section{Conclusion}

We introduce \benchmark, a lightweight and simulation-free testbed in textual embodied worlds. Designed to be fully embodied, \benchmark aims to test LLM agent's embodied planning capacity without errors from visual perceptions. To further address the embodiment challenge, we devise a novel CoT schema, \model{EmMem}, which summarizes embodied states \wrt history information. Although benchmark results indicate the effectiveness of \model{EmMem}, there still remains a huge performance gap compared with SFT results on individual tasks. As such, we call for future exploration of embodied generalists on \benchmark.

\section*{Limitation}
In this work, we endeavor to investigate how LLM performs and understands embodiment in a generic embodied textual setting. Nevertheless, our platform maintains the potential to be bridged with most simulators via additional visual language models (VLMs). Due to the page limitation and workload capacity, it remains unclear whether such linkage would be a successful attempt, even if we have obtained a powerful LLM embodied agent. We would like to leave it as a strong inclination for future research on generic embodied multimodal agents.

Moreover, the high demand of LLM API calls leads to a substantial querying price for experiments. Preliminary studies show that open-source language models, especially small models (\eg, LLAMA 2-7B), are unable to make appropriate embodied predictions in our environment, no matter what kind of prompting strategy is used.
Nevertheless, it would be more than valuable to investigate how open-source language models can be trained or finetuned in the \benchmark environment. 

Additionally, following previous works~\cite{yao2022react}, we mainly use GPT-4 and gpt-3.5-turbo for all experiments. Though we have observed performance boosts with a more powerful model, whether all observations and analyses are model-dependent remains unknown. Intuitively, our framework is model-agnostic and can be generalized to all embodied tasks. We are more than willing to experiment on more APIs once we have the access.

Lastly, we didn't go into too much on human communications and multi-agent tasks. Nevertheless, our platform maintains the capabilities to be generalized to multi-agent settings and human communicative settings (all you need is to update the configuration of agents). As such, learning and experimenting with multiple embodied agents in LangSuitE could also be a valuable future exploration trend

\section*{Ethics Statement}

In this work, we aim to deliver a generic, lightweight and simluation-free testbed for general embodied planning. By leveraging large language models, the main purpose of this research is to benefit the embodied AI and language reasoning community. Moreover, we devise a novel reasoning paradigm, \model{EmMem}, which grants agents with emergent embodied memories. Going beyond embodied understanding, we believe the CoT paradigm is general and can be generalized to more reasoning and planning tasks.

\section*{Acknowledgements}
The authors thank Dr. Wentao Ding for his great contributions in refining code repository and Mrs. Chen Zhen for her design of illustrative teasers. This work presented herein is supported by the National Natural Science Foundation of China (62376031). 

{
    \bibliography{main}
}

\input{X_Appendix}

\end{document}

%% file: config.tex
\usepackage{enumitem}
\usepackage{graphicx}
\graphicspath{figures/}
\usepackage{multirow}
\usepackage{acronym}
\usepackage{xspace}
\usepackage{wrapfig}
\usepackage{booktabs}
\usepackage{titletoc}
\usepackage{appendix}
\usepackage{pifont}
\usepackage[tableposition=top]{caption}

\usepackage[linesnumbered,ruled,vlined]{algorithm2e}
\usepackage{bbm}
\usepackage{float}
\usepackage{colortbl}

\usepackage{makecell}

\usepackage{diagbox}
\usepackage{verbatim}
\usepackage{framed}
\usepackage[most]{tcolorbox}

\usepackage[capitalize]{cleveref}
\usepackage{xcolor}

\definecolor{myblue}{RGB}{151,198,194}
\definecolor{mygreen}{RGB}{181,209,177}
\definecolor{myred}{RGB}{147,44,80}
\definecolor{gb}{RGB}{68,179,110}
\definecolor{rb}{RGB}{147,44,80}

\makeatletter
\DeclareRobustCommand\onedot{\futurelet\@let@token\@onedot}
\def\@onedot{\ifx\@let@token.\else.\null\fi\xspace}
\def\eg{\emph{e.g}\onedot} 

\def\ie{\emph{i.e}\onedot}

\def\etc{\emph{etc}\onedot} 

\def\wrt{w.r.t\onedot}

\makeatother

\makeatletter
\renewcommand{\paragraph}{%
  \@startsection{paragraph}{4}%
  {\z@}{0ex \@plus 0ex \@minus 0ex}{-1em}%
  {\normalfont\normalsize\bfseries}%
}
\makeatother

\usepackage{listings}

\definecolor{codegreen}{rgb}{0,0.6,0}

\DeclareCaptionFont{white}{\color{white}}
\DeclareCaptionFormat{listing}
  {\colorbox{gray}
     {\parbox{\dimexpr\textwidth-2\fboxsep}{\centering #1#2#3}}}
\acrodef{llm}[LLM]{Large Language Model}
\acrodef{kg}[KG]{knowledge graph}
\acrodef{rl}[RL]{reinforcement learning}
\acrodef{cot}[CoT]{chain-of-thought}

%% file: X_Appendix.tex
\clearpage
\appendix

\section{The motivation of the design for \benchmark}
First, we want to design our unified textual environment being general, embodied, interactive, and communicative, thereby containing most of the typical embodied task categories, such as navigation, rearrangement, human interaction, multi-agent cooperation, and so on. Therefore, we transfer typical embodied tasks, carefully selecting action sets that are both suitable for our environment and the original task. The differences between our benchmark and original embodied tasks are that our environment is fully embodied and the actions are low-level (fine-grained atom actions). 

Second, As the previous platform generally gives the global description of the environment, trained models in the simulated environment are difficult to apply to real environments. Our platform simulation more realistic situation that agent has a limited view scope and action scope.

Third,  The simulation-free platform makes various embodied tasks conveniently and easily performed in a unified fashion.

\section{Data and Code Release}

All code and data will be publicly available for use. The data and code are under MIT license. All resources can only be used for research purposes. More details of the construction process of \benchmark can be found in the codebase.

\section{Time complexity comparison}
We compare the computational cost for one-step execution/rendering on simulated environments vs. LangSuit·E. The time interval (\textbf{Time Int}) counts from receiving one LLM inference results to the next query to LLM (thus the latency caused by API-call is not counted). We run all experiments on an Intel Core i9 with 64GB RAM. Results are shown in Tab. \ref{tab: speed}. Our \benchmark demonstrates significant systemwide efficiency improvements.

\begin{table}
    \centering
    \small
    \caption{Time complexity comparison of simulated environments with \benchmark. \textbf{Omni} is the abbreviation of OmniGibson (OmniVerse) which is a physical-based simulator, and \textbf{AI2Thor} (Unity) takes only symbolic actions.}
    \label{tab: speed}

    \begin{tabular}{l|ccc}
    \toprule
       \textbf{Env. (Engine)} & \textbf{Omni} & \textbf{AI2Thor} & \textbf{\benchmark}\\
        
        \midrule
        \prompt{\textbf{Time Int (ms)}}  & 67.90 & 7.81 & 1.89 \\
    \bottomrule
    \end{tabular}
\end{table}

\begin{table}[t!]
    \centering
    \small
    \caption{\textbf{Evaluation results on  Household task with Path Weighted Metrics.}}
    \begin{tabular}{l|cc}
    \toprule
        \multirow{2}{*}{\textbf{Model}} & \multicolumn{2}{c}{\textbf{Household}}  \\

        & \textbf{SR.} &  \textbf{Goal-SR.}   \\
        
        \midrule
       \textit{Low-Level Actions Setting}
       \\\midrule
        \prompt{Act}  & 6.65 & 21.94 \\
        \prompt{ReAct}  & 9.50 & 24.71\\
        \prompt{ReAct-EmMem}  & 7.19 & 21.13 \\
        \prompt{Reflexion}  & 14.89 & 25.73 \\
        \prompt{Reflexion-EmMem}  & 11.19 & 22.77 \\
        \midrule
       \textit{High-Level Actions Setting}
       \\\midrule
        \prompt{Act} & 22.38 & 37.13  \\
        \prompt{ReAct} & 18.64 & 35.52  \\
        \prompt{ReAct-EmMem} & 28.27 & 43.02 \\
        \prompt{Reflexion}  & 28.23 & 40.00  \\
        \prompt{Reflexion-EmMem}  & 34.45 & 43.80 \\
    \bottomrule
    \end{tabular}
\label{tab:SPL}
\end{table}

\section{Path Weighted Metrics}
\label{app:path}
We additionally utilized Path Weighted Metrics \cite{shridhar2020alfred} to calculate the path weighted score on Household task. This weighted score penalizes results that complete the task but take a longer number of steps. The introduction of Path Weighted Metrics allows the evaluation to focus not only on task completion but also on efficiency and path length, thereby providing a more comprehensive assessment of the model's performance. The path weighted score $p_s$ for metric $s$ is given as:
\begin{equation}
    p_s = s \times \frac{L^*}{max(L^{*}, \hat{L})}
\end{equation}
where $\hat{L}$ is the number of actions the agent took in the episode, and $L^*$ is the number of actions in the expert demonstration.

As can be seen from the Tab. \ref{tab:SPL}, after using the path weighted score, the performance on the \textbf{Act} strategy does not decline significantly. However, for other tasks, the agent is encouraged to think more, leading to an increase in the number of steps and consequently greater penalties, resulting in a more noticeable decline.

\section{Customized configuration of an agent.}
LangSuit·E supports the customized configuration of agents. It allows the configuration of prompt strategies, the initialization position of agents, inventory capacity, focal length, manipulate distance and view distance. For more details, see Tab. \ref{tab:agent_config}.
\begin{table}[h!]
\small

\caption{Customized configuration of an agent.}
\centering
\label{tab:agent_config}
\resizebox{\linewidth}{!}{%
\begin{tabular}{l|p{5cm}}
\toprule
\textbf{Type} & \textbf{Description} \\ \midrule
\prompt{agent\_type} &  Whether an agent has ReAct ability.\\
\prompt{agent\_position} &  The initial position of an agent. \\
\prompt{inventory\_capacity} &  The number of objects an agent can hold.\\
\prompt{focal\_length} &  Inversely proportional to viewing angle; shorter focal length corresponding to larger view field.  \\
\prompt{manipulate\_distance} & The max distance that an agent's hand can reach. \\
\prompt{view\_distance} &  The max distance that an agent's view can reach.\\
\bottomrule
\end{tabular}
}

\end{table}

\begin{table}[h!]
\small
\centering
\caption{\textbf{Action settings in \benchmark.}}

\resizebox{\linewidth}{!}{%
\begin{tabular}{l|p{5cm}}
\toprule
\textbf{Action Type} & \textbf{Description} \\ \midrule
\multicolumn{2}{l}{\textit{Low-level actions}} \\ \midrule
\prompt{move\_ahead} & Move ahead by 1 step. \\
\prompt{turn\_left} & Turn left by 90 degrees. \\
\prompt{turn\_right} & Turn right by 90 degrees. \\
\prompt{pan\_left} &  Move left by 1 step. \\
\prompt{pan\_right} & Move right by 1 step. \\
\prompt{pick\_up [obj]} & Pick up an object identified by \prompt{[obj]}. \\
\prompt{drop [obj]} & Drop the object \prompt{[obj]} to the agent's current position. \\ 
\prompt{put [obj, recep]} & Put the object \prompt{[obj]} into the receptacle \prompt{[recep]}. \\ 
\prompt{open [obj]} & Open object \prompt{[obj]} if it is openable. \\
\prompt{close [obj]} & close object \prompt{[obj]} if it is closeable. \\
\prompt{toggle\_on [obj]} & Toggle on object \prompt{[obj]} if it is toggleable. \\
\prompt{toggle\_off [obj]} & Toggle off object \prompt{[obj]} if it is toggleable. \\
\prompt{slice [obj]} & Slice object \prompt{[obj]} if it is sliceable. \\
\prompt{stop [answer]} & Stop making actions and conclude the final answer in \prompt{[answer]}.\\ 
\midrule
\multicolumn{2}{l}{\textit{High-level actions}} \\ \midrule
\prompt{goto [loc]} & Go to a location that is in current observation. \\ 
\prompt{heat [obj, heater]} & Heat object with a microwave or a stove burner. \\ 
\prompt{cool [obj, cooler]} & Cool object with a fridge. \\ 
\prompt{clean [obj, sink]} & Clean object with a basin, a sink or a bathtub. \\
\prompt{go\_explore [room]} & Explore around the room.\\
\prompt{go\_check [recep]} & Go to receptacle and check its inside.\\
\prompt{go\_grab [obj]} & Go to object and pick it up.\\
\prompt{go\_put [recep]} & Go to receptacle and put the object in hand into it.\\
\midrule
\multicolumn{2}{l}{\textit{Communicative actions}} \\ \midrule
\prompt{chat [message]} & Send messages to other agents. \\
\prompt{ask [message]} & Send queries to users.\\
\bottomrule
\end{tabular}
}
\label{tab: complete_action_type}
\end{table}

\section{Pseudo-Code for Expert Trajectory Generation}

\cref{alg: exp_gen} shows the pseudo-code for generating expert trajectories as references.

\begin{algorithm}[h!]
    \DontPrintSemicolon
    \SetNoFillComment
    \SetKwFunction{CompareStatus}{CompareStatus}
    \SetKwFunction{SubtaskPlaner}{SubtaskPlaner}
    \SetKwFunction{AStarPathPlaner}{AStarPathPlaner}
    \SetKwData{StatusDiff}{statusDiff}
    \SetKwData{Plan}{plan}
    \SetKwData{Subtask}{subtask}
    \SetKwData{AgentPosition}{agentPosition}
    \SetKwData{ObjectPosition}{objectPosition}
    \SetKwData{MoveActions}{moveActions}
    \SetKwData{OperationActions}{operationActions}
    \SetKwData{FinalActions}{finalActions}
    \caption{Expert Trajectory Generation}
    \label{alg: exp_gen}

        \KwIn {Initialized environment $E$, start status of environment  $\mathcal{S}=\{\mathbf{s}_i\}$, target status of environment  $\mathcal{T}=\{\mathbf{t}_i\}$}
        \KwOut {a list of actions}
        \StatusDiff $\leftarrow$ compare start status and target status by function \;
        \CompareStatus($\mathcal{S}$, $\mathcal{T}$)\;
        \Plan $\leftarrow$ obtain a sequence of sub-tasks by \SubtaskPlaner(StatusDiff)\;
        \While{\Plan $\neq \emptyset$}{
        \Subtask $\leftarrow$ pop a sub-task from \Plan.\;
        \AgentPosition $\leftarrow$ get current agent position form environment $E$\;
        \ObjectPosition $\leftarrow$ get object position form \Subtask\;
        \MoveActions $\leftarrow$ compute shortest path by  \AStarPathPlaner(\AgentPosition, \ObjectPosition)\;
        \OperationActions $\leftarrow$ get Operation Actions form \Subtask\;
        \For{ $action$ $\in$ \MoveActions }{
        Update \FinalActions by append the $action$ into $\FinalActions$\;
        Update environment $E$ by performing $action$\;
        }\;
        \For{ $action$ $\in$ \OperationActions }{
        Update \FinalActions by append the $action$ into \FinalActions\;      
        Update environment $E$ by performing $action$\;
        }
        }
    \Return {\FinalActions} \;
\end{algorithm}

\section{Examples of 5 prompt strategies for task Instruction Grounding}
\label{app:prompt}
We show the detailed instruction for different prompt strategies: \textbf{Act}, \textbf{ReAct}, \textbf{ReAct+\model{EmMem}}, \textbf{Reflexion}, and \textbf{Reflexion+\model{EmMem}}. We highlight the pivotal instruction sentences.
\subsection{Act}
\begin{lstlisting}
Instruction: 
You are an autonomous intelligent agent tasked with completing a minigrid game. These tasks will be accomplished through the use of specific actions you can issue. 

Here's the information you'll have: 
Task: This is the task you are trying to accomplish.
Obs: These are the objects you can see.You can see at most {max_view_steps} step(s) in front of you, {side_steps} step(s) on your left, {side_steps} step(s) on your right. 
Manipulable object: The object that you are facing and you can take an operate action like pick_up [object_name], toggle [object_name] on it. 
Inventory: These are the objects you are holding, and you can take an action like drop on it.
Feedback: Whether the action is succeed or not and why is it.At each step, you can choose to think your plan or execute an action from one of the following: 
`move_ahead`: it means you move ahead by 1 step.
`turn_left`: it means you turn left 90 degrees.
`turn_right`: it means you turn right 90 degrees.`pick_up [object_name]`: Use this action to pick up a pickupable and visible object directly in front of you without obstacle.`drop [object_name]`: it means you places the object in hand at the place just in front of you.`toggle [object_name]`: it means you toggle a toggleable and visible object directly in front of you without obstacle.
`stop [answer]`: Issue this action when you believe the task is complete. If you believe the task is impossible to complete, provide the answer as \"N/A\" in the bracket.

To be successful, it is very important to follow the following rules: 
1. You should only issue a action that is valid given the current observation
2. You should only issue ONE action at a time, and ONLY action is enough.
3. Generate the action in the correct format starting with \"Act: \". For example, \"Act: pick_up [ball]\".
4. You can only operate the object you are facing, if the object is in front of you, on your left or right, take action like move_ahead, turn_left, turn_right to appraoch it until you are facing it;
5. Issue stop action when you think you have achieved the objective. Don't generate anything after stop.

Here are some examples: 
{example}
Respond YES if you can play this game.

Example: 
Task: go to the red box. 
Obs: You can see a blue key in front of you; You can see a red box on your right. 
Manipulable object: A blue key.
>Act: turn_right.

Task: 
Task: {task}
Obs: {observation}
What is your next step?

\end{lstlisting}

\subsection{ReAct}
\begin{lstlisting}
Instruction: 
You are an autonomous intelligent agent tasked with completing a minigrid game. These tasks will be accomplished through the use of specific actions you can issue. 

Here's the information you'll have: 
Task: This is the task you are trying to accomplish.
Obs: These are the objects you can see.You can see at most {max_view_steps} step(s) in front of you, {side_steps} step(s) on your left, {side_steps} step(s) on your right. 
Manipulable object: The object that you are facing and you can take an operate action like pick_up [object_name], toggle [object_name] on it. 
Inventory: These are the objects you are holding, and you can take an action like drop on it.
Feedback: Whether the action is succeed or not and why is it.

At each step, you can choose to think your plan or execute an action from one of the following: 
`move_ahead`: it means you move ahead by 1 step.
`turn_left`: it means you turn left 90 degrees.
`turn_right`: it means you turn right 90 degrees.
`pick_up [object_name]`: Use this action to pick up a pickupable and visible object directly in front of you without obstacle.
`drop [object_name]`: it means you places the object in hand at the place just in front of you.
`toggle [object_name]`: it means you toggle a toggleable and visible object directly in front of you without obstacle.
`stop [answer]`: Issue this action when you believe the task is complete. If you believe the task is impossible to complete, provide the answer as \"N/A\" in the bracket.

To be successful, it is very important to follow the following rules: 
1. You should only issue a action or your thought that is valid given the current observation
2. You should only issue ONE action at a time, and ONLY action is enough.
3. Generate the action in the correct format starting with \"Act: \". For example, \"Act: pick_up [cup_0]\".
4. State your thought or think process starting with \"Thought: \"
5. You can only operate the object you are facing, if the object is in front of you, on your left or right, take action like move_ahead, turn_left, turn_right to appraoch it until you are facing it;
6. Issue stop action when you think you have achieved the objective. Don't generate anything after stop.

Here are some examples: 
{example}
Respond YES if you can play this game.

Example: 
Task: go to the red box. 
Obs: You can see a blue key in front of you; You can see a red box on your right. 
Manipulable object: A blue key.
>Thought: To solve this task, I need to find a red box. There is a blue key just before me and blocked my way. The red box is on my right, so I should turn\_right first to go to the red box.  

>Act: turn_right.

Task: 
Task: {task}
Obs: {observation}
What is your next step? Try to think before act.

\end{lstlisting}

\subsection{ReAct+\model{EmMem}}
\begin{lstlisting}
Instruction: 
You are an autonomous intelligent agent tasked with completing a minigrid game. These tasks will be accomplished through the use of specific actions you can issue. 

Here's the information you'll have: 
Task: This is the task you are trying to accomplish.
Obs: These are the objects you can see.You can see at most {max_view_steps} step(s) in front of you, {side_steps} step(s) on your left, {side_steps} step(s) on your right. 
Manipulable object: The object that you are facing and you can take an operate action like pick_up [object_name], toggle [object_name] on it. 
Inventory: These are the objects you are holding, and you can take an action like drop on it.
>Feedback: Whether the action is succeed or not and why is it.

At each step, you can choose to think your plan or execute an action from one of the following: 
`move_ahead`: it means you move ahead by 1 step.
`turn_left`: it means you turn left 90 degrees.
`turn_right`: it means you turn right 90 degrees.
`pick_up [object_name]`: Use this action to pick up a pickupable and visible object directly in front of you without obstacle.
`drop [object_name]`: it means you places the object in hand at the place just in front of you.
`toggle [object_name]`: it means you toggle a toggleable and visible object directly in front of you without obstacle.
`stop [answer]`: Issue this action when you believe the task is complete. If you believe the task is impossible to complete, provide the answer as \"N/A\" in the bracket.

To be successful, it is very important to follow the following rules: 
1. You should only issue a action or your thought that is valid given the current observation
2. You should only issue ONE action at a time, and ONLY action is enough.
3. Generate the action in the correct format starting with \"Act: \". For example, \"Act: pick_up [cup_0]\".
4. State your thought or think process starting with \"Thought: \"
5. You can only operate the object in your sight, if the object is in front of you, on your left or right, take action like move_ahead, turn_left, turn_right to appraoch it until you it in your sight;
6. Issue stop action when you think you have achieved the objective.\n7. Remember you are facing NORTH at the begining.
Don't generate anything after stop.

Here are some examples: 
{example}
Respond YES if you can play this game.

Example: 
Task: go to the red box. 
Obs: You can see nothing ahead.
>Thought: I see nothing, it could be I have reached the border or I am in a middle grid but there is nothing in front of me. I have taken 0 steps and I am facing NORTH now. I should remember what I saw, so that I can find them easily when necessary; I need to find a red box. 
>OK.
>Act: move_ahead 
>Feedback: Action failed. Can not move ahead, because there is an obstacle ahead.
Obs: You see nothing ahead.
>Thought: I can't move ahead because of an obstacle but I saw nothing in front of me. so I have reached the border, I have taken 0 step and am facing NORTH now since last action failed. Next I will try to look around. 
>OK.
>Act: turn_right 
>Feedback: Action succeeded. Turned right by '90' degrees.
Obs: You see a red box.
>Thought: I have reached the border when facing NORTH. I haven taken 2 steps and am facing EAST after turn right at last step. The border is on my leftside. Now I find the red box. Next I need to go to it. 
>OK.
>Act: move_ahead
>Feedback: Action succeeded. Moved forward by 1 step.
Obs: You can see a red box.
>Thought: I am moving along the border now, I have taken 3 steps and am still facing EAST after move ahead at last step. I need to go to the red box. 
>OK.
>Act: move_ahead()
>Feedback: Action failed. Can not move ahead, because there is an obstacle ahead. 
Obs: You can see a red box.
>Thought: I see a red box and can not move ahead, I have taken 3 steps and am still facing EAST since last action failed. I have reached the red box. 
>OK.
>Act: stop[].

Task: 
Task: {task}
Obs: {observation}
What is your next step? Try to summarize your status, recall what you have done and think before act. 
\end{lstlisting}

\subsection{Reflexion}
\begin{lstlisting}
Instruction: 
You are an autonomous intelligent agent tasked with completing a minigrid game. These tasks will be accomplished through the use of specific actions you can issue. 

Here's the information you'll have: 
Task: This is the task you are trying to accomplish.
Obs: These are the objects you can see.You can see at most {max_view_steps} step(s) in front of you, {side_steps} step(s) on your left, {side_steps} step(s) on your right.
Manipulable object: The object that you are facing and you can take an operate action like pick_up [object_name], toggle [object_name] on it. 
Inventory: These are the objects you are holding, and you can take an action like drop on it.
Feedback: Whether the action is succeed or not and why is it.

 At each step, you can choose to think your plan or execute an action from one of the following: 
`move_ahead`: it means you move ahead by 1 step.
`turn_left`: it means you turn left 90 degrees.
`turn_right`: it means you turn right 90 degrees.
`pick_up [object_name]`: Use this action to pick up a pickupable and visible object directly in front of you without obstacle.
`drop [object_name]`: it means you places the object in hand at the place just in front of you.
`toggle [object_name]`: it means you toggle a toggleable and visible object directly in front of you without obstacle.
`stop [answer]`: Issue this action when you believe the task is complete. If you believe the task is impossible to complete, provide the answer as \"N/A\" in the bracket.

To be successful, it is very important to follow the following rules:
1. You should only issue a action or your thought that is valid given the current observation
2. You should only issue ONE action at a time, and ONLY action is enough.
3. Generate the action in the correct format starting with \"Act: \". For example, \"Act: pick_up [cup_0]\".
4. State your thought or think process starting with \"Thought: \"
5. You can only operate the object you are facing, if the object is in front of you, on your left or right, take action like move_ahead, turn_left, turn_right to appraoch it until you are facing it;
6. Issue stop action when you think you have achieved the objective. Don't generate anything after stop.

Here are some examples: 
{example}
Respond YES if you can play this game.

Example: 
Task: go to the red box. 
Obs: You can see a blue key in front of you; You can see a red box on your right. 
Manipulable object: A blue key.
>Thought: To solve this task, I need to find a red box. There is a blue key just before me and blocked my way. The red box is on my right, so I should turn_right first to go to the red box.
>Act: turn_right.

Task: 
Task: {task}
Your memory from last trails is: \{memory\} 
Obs: {observation}
What is your next step? Try to think before act.
\end{lstlisting}

\subsection{Reflexion+\model{EmMem}}
\begin{lstlisting}
Instruction: 
You are an autonomous intelligent agent tasked with completing a minigrid game. These tasks will be accomplished through the use of specific actions you can issue. 

Here's the information you'll have: 
Task: This is the task you are trying to accomplish.
Obs: These are the objects you can see.You can see at most {max_view_steps} step(s) in front of you, {side_steps} step(s) on your left, {side_steps} step(s) on your right. 
Manipulable object: The object that you are facing and you can take an operate action like pick_up [object_name], toggle [object_name] on it. 
Inventory: These are the objects you are holding, and you can take an action like drop on it.
>Feedback: Whether the action is succeed or not and why is it.

At each step, you can choose to think your plan or execute an action from one of the following: 
`move_ahead`: it means you move ahead by 1 step.
`turn_left`: it means you turn left 90 degrees.
`turn_right`: it means you turn right 90 degrees.
`pick_up [object_name]`: Use this action to pick up a pickupable and visible object directly in front of you without obstacle.
`drop [object_name]`: it means you places the object in hand at the place just in front of you.
`toggle [object_name]`: it means you toggle a toggleable and visible object directly in front of you without obstacle.
`stop [answer]`: Issue this action when you believe the task is complete. If you believe the task is impossible to complete, provide the answer as \"N/A\" in the bracket.

To be successful, it is very important to follow the following rules: 
1. You should only issue a action or your thought that is valid given the current observation
2. You should only issue ONE action at a time, and ONLY action is enough.
3. Generate the action in the correct format starting with \"Act: \". For example, \"Act: pick_up [cup_0]\".
4. State your thought or think process starting with \"Thought: \"
5. You can only operate the object in your sight, if the object is in front of you, on your left or right, take action like move_ahead, turn_left, turn_right to appraoch it until you it in your sight;
6. Issue stop action when you think you have achieved the objective.\n7. Remember you are facing NORTH at the begining.
Don't generate anything after stop.

Here are some examples: 
{example}
Respond YES if you can play this game.

Example: 
Task: go to the red box. 
Obs: You can see nothing ahead.
>Thought: I see nothing, it could be I have reached the border or I am in a middle grid but there is nothing in front of me. I have taken 0 steps and I am facing NORTH now. I should remember what I saw, so that I can find them easily when necessary; I need to find a red box.
>OK.
>Act: move_ahead 
>Feedback: Action failed. Can not move ahead, because there is an obstacle ahead.
Obs: You see nothing ahead.
>Thought: I can't move ahead because of an obstacle but I saw nothing in front of me. so I have reached the border, I have taken 0 step and am facing NORTH now since last action failed. Next I will try to look around.
>OK.
>Act: turn_right 
>Feedback: Action succeeded. Turned right by '90' degrees.
Obs: You see a red box.
>Thought: I have reached the border when facing NORTH. I haven taken 2 steps and am facing EAST after turn right at last step. The border is on my leftside. Now I find the red box. Next I need to go to it.
>OK.
>Act: move_ahead
>Feedback: Action succeeded. Moved forward by 1 step.
Obs: You can see a red box.
>Thought: I am moving along the border now, I have taken 3 steps and am still facing EAST after move ahead at last step. I need to go to the red box.
>OK.
>Act: move_ahead()
>Feedback: Action failed. Can not move ahead, because there is an obstacle ahead. 
Obs: You can see a red box.
>Thought: I see a red box and can not move ahead, I have taken 3 steps and am still facing EAST since last action failed. I have reached the red box.
>OK.
>Act: stop[].

Task: 
Task: {task}
Your memory from last trails is: {memory}
Obs: {observation}
What is your next step? Try to summarize your status, recall what you have done and think before act. 
\end{lstlisting}

\section{Examples of different task prompts}
\subsection{Instruction Grounding}

\begin{lstlisting}
System: You are an autonomous intelligent agent tasked with completing a minigrid game. These tasks will be accomplished through the use of specific actions you can issue. 

Here's the information you'll have: 
Task: This is the task you are trying to accomplish.
Obs: These are the objects you can see.You can see at most 7 step(s) in front of you, 3 step(s) on your left, 3 step(s) on your right. 
 Manipulable object: The object that you are facing and you can take an operate action like pick_up [object_name], toggle [object_name] on it. 
Inventory: These are the objects you are holding, and you can take an action like drop on it.
Feedback: Whether the action is succeed or not and why is it.

At each step, you can choose to think your plan or execute an action from one of the following: 
`move_ahead`: it means you move ahead by 1 step.
`turn_left`: it means you turn left 90 degrees.
`turn_right`: it means you turn right 90 degrees.
`pick_up [object_name]`: Use this action to pick up a pickupable and visible object directly in front of you without obstacle.
`drop [object_name]`: it means you places the object in hand at the place just in front of you.
`toggle [object_name]`: it means you toggle a toggleable and visible object directly in front of you without obstacle.
`stop [answer]`: Issue this action when you believe the task is complete. If you believe the task is impossible to complete, provide the answer as "N/A" in the bracket.

To be successful, it is very important to follow the following rules: 
1. You should only issue a action or your thought that is valid given the current observation
2. You should only issue ONE action at a time, and ONLY action is enough.
3. Generate the action in the correct format starting with "Act: ". For example, "Act: pick_up [cup_0]".
4. State your thought or think process starting with "Thought: "
5. You can only operate the object you are facing, if the object is in front of you, on your left or right, take action like move_ahead, turn_left, turn_right to appraoch it until you are facing it;
6. Issue stop action when you think you have achieved the objective. Don't generate anything after stop.

Here are some examples: 
Task: go to the red box. 
Obs: You can see a blue key in front of you; You can see a red box on your right. 
Manipulable object: A blue key.
>Thought: To solve this task, I need to find a red box. There is a blue key just before me and blocked my way. The red box is on my right, so I should turn_right first to go to the red box.
>Act: turn_right.
Respond YES if you can play this game.
Assistant: YES
System: Task: pick up the red key
Obs: You can see a yellow door, a yellow key in front of you; You can see a green door on your left; You can see a green box,a red key on your right.
What is your next step? 
Assistant: Thought:  To accomplish the task, I need to pick up the red key. The red key is on my right, so the first step is to turn right.
System: OK.
Assistant: Act: turn_right.
System: Feedback: Action succeeded. Turned right by '90' degrees.
Obs: You can see a red key in front of you; You can see a green door,a blue key,a green box,a yellow key on your left; You can see a purple ball,a purple door on your right.
Assistant: Thought:  Now that I'm facing the red key, I can pick it up to accomplish my task.
System: OK.
Assistant: Act: pick_up [red key].
System: Feedback: Action failed. Failed to pick up red key. You can only pickup the object one step in front of you without obstacle.
Obs: You can see a red key in front of you; You can see a green door,a blue key,a green box,a yellow key on your left; You can see a purple ball,a purple door on your right.
Assistant: Thought: The red key is still in front of me, but I may need to move closer to pick it up.
System: OK.
Assistant: Act: move_ahead.
System: Feedback: Action succeeded. Moved forward by 1 step.
Obs: You can see a red key in front of you; You can see a green door,a blue key,a green box on your left; You can see a purple ball on your right.
Assistant: Thought:  The red key is now directly in front of me. I can pick it up now.
System: OK.
Assistant: Act: pick_up [red key].
System: Feedback: Action failed. Failed to pick up red key. You can only pickup the object one step in front of you without obstacle.
Obs: You can see a red key in front of you; You can see a green door,a blue key,a green box on your left; You can see a purple ball on your right.
Assistant: Thought:  I must have been too far to pick up the red key. I should move one step ahead to reach the key.
System: OK.
Assistant: Act: move_ahead.
System: Feedback: Action succeeded. Moved forward by 1 step.
Obs: You can see a red key in front of you; You can see a green door,a blue key on your left; 
Manipulable object: a red key.
Assistant: Thought:  I am now facing the red key and can pick it up.
System: OK.
Assistant: Act: pick_up [red key].
System: Feedback: Action succeeded. You picked a red key up.
Inventory: You are now holding a red key.
Obs: You can see a green door,a blue key on your left.
System: [SUCCESS] You have completed the task. Congratulations!
System: DONE!
\end{lstlisting}

\subsection{Rearrangement}
\begin{lstlisting}
System: You are an autonomous intelligent agent tasked with rearranging a vitual home. System will randomly change between 1 to 5 objects in the room. Your goal is to identify which objects have changed and reset those objects to their original state.
Here are the information you will have: 
Original status: The original room status that you should remember, pay your attention to the position and openness states of each of the objects. Then find out all the changed objects and reset them to their origianl states.
Obs: These are the objects you can see. You can see at most 8.0 step(s) in front of you; 60 degrees on your left, 60 degrees on your right.
Inventory: These are the objects you are holding. You can hold at most 1 objects.
Feedback: Whether the action is succeed or not and why is it.

 At each step, you can choose to think your plan or execute an action from one of the following: 
`move_ahead`: it means you move ahead by 1 step.
`turn_left`: it means you turn left 90 degrees.
`turn_right`: it means you turn right 90 degrees.
`pick_up [object_name]`: Use this action to pick up a pickupable and visible object within 8.0 step(s).
`drop [object_name]`: it means you place the object in hand at the place just in front of you.
`open [object_name]`: it means you open an openable and visible object within 8.0 step(s).
`close [object_name]`: it means you close a closeable and visible object.
`stop [answer]`: Issue this action when you believe the task is complete. If you believe the task is impossible to complete, provide the answer as 'N/A' in the bracket.

To be successful, it is very important to follow the following rules: 
1. You should only issue ONE action or ONE thought that is valid given the current observation
2. When you issue a action, and ONLY ONE action is enough, and then wait for your feedback from system.
3. Generate the action in the correct format starting with "Act: " For example, "Act: pick_up [cup_0]" Do not add adjectives before object name.
4. State your thought or think process starting with "Thought: "
5. You can ONLY operate the object within your view distance.
6. If got action failed feedback, try to take other actions to proceed the task. 
7. Issue stop action when you think you have achieved the objective. Don't generate anything after stop.

Here are some examples: 
Original status: In the north of the room, there is a armchair_0; a pen_1; In the east of the room, there is a chair_0;.
Obs: You can see nothing ahead.
> Thought: To solve the task, I need to explore the room to find out what objects have been changed. I see a pen_1, because I am in the south of the room, so the pen_1 is in the south of the room, but I remember the pen_1 should be in the north of the room according to the origianl status. So I need to pick it up and put it back.
OK.
> Act: move_ahead

Respond YES if you can play this game.
Rearrange Agent: YES
System: Let's start.
Original state: In the north of the room, there is a cabinet_2; a cabinet_3; a cabinet_13; a cabinet_16; a cabinet_19; a cabinet_22; a microwave_0; a sink_0; a sink_1, in/on it you can see a dishsponge_0. In the northeast of the room, there is a lightswitch_0. In the east of the room, there is a cabinet_7; a cabinet_20; a countertop_0, in/on it you can see a coffeemachine_0, a fork_0, a papertowelroll_0, a peppershaker_0, a saltshaker_0, a spatula_0, a spoon_0; a drawer_0; a drawer_1; a garbagecan_0; a stoveknob_2. In the southeast of the room, there is a cabinet_17; a countertop_1, in/on it you can see a cup_0, a plate_0; a drawer_2; a stoveburner_0; a stoveburner_1; a stoveburner_2, in/on it you can see a pot_0; a stoveburner_3; a stoveknob_0; a stoveknob_1; a stoveknob_3. In the south of the room, there is a diningtable_0, in/on it you can see a apple_0, a bread_0, a lettuce_0, a mug_0, a potato_0, a tomato_0; a stool_0; a window_0. In the southwest of the room, there is a cabinet_5; a cabinet_12; a cabinet_14; a cabinet_21. In the west of the room, there is a cabinet_0; a cabinet_4; a cabinet_9; a cabinet_11; a fridge_0, in/on it you can see a egg_0. In the northwest of the room, there is a cabinet_1; a cabinet_6; a cabinet_8; a cabinet_10; a cabinet_15; a cabinet_18; a countertop_2, in/on it you can see a bowl_0, a butterknife_0, a faucet_0, a faucet_1, a knife_0, a pan_0, a soapbottle_0, a toaster_0; a drawer_3. 
Obs: In front of you, You see a cabinet_0; a cabinet_4; a cabinet_9; a cabinet_11; an opened fridge_0, there is a egg_0 in/on it. On your left, you see a cabinet_2; a cabinet_10; a cabinet_13; a cabinet_18; a countertop_2; a bowl_0; a faucet_1; a knife_0; a pan_0; a toaster_0; a diningtable_0; a apple_0; a bread_0; a butterknife_0; a dishsponge_0; a lettuce_0; a potato_0; a saltshaker_0; a soapbottle_0; a spoon_0; a tomato_0; a drawer_3; a microwave_0; a sink_0; a sink_1; a stool_0; a window_0. On your right, you see a cabinet_3; a cabinet_22; a faucet_0. 
what is your next step? 
Assistant: Thought:  As there are many objects in the room, I need to identify which objects have changed. Since I can see an egg in the fridge but the fridge is open, I will close the fridge to reset it to its original state.
Assistant: Act: close fridge_0.
System: Feedback: Action succeeded. You closed fridge_0.
Obs: In front of you, You see a cabinet_0; a cabinet_4; a cabinet_9; a cabinet_11; a fridge_0. On your left, you see a cabinet_2; a cabinet_10; a cabinet_13; a cabinet_18; a countertop_2; a bowl_0; a faucet_1; a knife_0; a pan_0; a toaster_0; a diningtable_0; a apple_0; a bread_0; a butterknife_0; a dishsponge_0; a lettuce_0; a potato_0; a saltshaker_0; a soapbottle_0; a spoon_0; a tomato_0; a drawer_3; a microwave_0; a sink_0; a sink_1; a stool_0; a window_0. On your right, you see a cabinet_3; a cabinet_22; a faucet_0. 
Assistant: Thought:  Now that the fridge is closed, I can see the original status of the room. I need to check if there are any changes in the room.
Assistant: Act: turn_left.
System: Feedback: Action succeeded. Turned left by '90' degrees.
Obs: In front of you, You see a diningtable_0; a apple_0; a dishsponge_0; a spoon_0; a window_0. On your left, you see a bread_0; a butterknife_0; a lettuce_0; a potato_0; a saltshaker_0; a soapbottle_0; a tomato_0; a stool_0; a stoveburner_0; a pot_0; a stoveburner_3. On your right, you see a cabinet_17; a countertop_1; a cup_0; a plate_0; a drawer_2; a fridge_0; a stoveburner_1; a stoveburner_2; a stoveknob_0; a stoveknob_3. 
Assistant: Thought:  I can see a dishsponge in the dining table which was not there before. So, I need to pick it up and put it back to its original position.
Assistant: Act: pick_up dishsponge_0.
System: Feedback: Action succeeded. You picked dishsponge_0 up.
Inventory: dishsponge_0.
Obs: In front of you, You see a diningtable_0; a apple_0; a spoon_0; a window_0. On your left, you see a bread_0; a butterknife_0; a lettuce_0; a potato_0; a saltshaker_0; a soapbottle_0; a tomato_0; a stool_0; a stoveburner_0; a pot_0; a stoveburner_3. On your right, you see a cabinet_17; a countertop_1; a cup_0; a plate_0; a drawer_2; a fridge_0; a stoveburner_1; a stoveburner_2; a stoveknob_0; a stoveknob_3. 
Assistant: Thought:  I need to put the dishsponge back to the sink_0 where it was originally placed.
Assistant: Act: move_ahead.
System: Feedback: Action succeeded. Moved forward by 1 step.
Obs: In front of you, You see a diningtable_0; a window_0. On your left, you see a apple_0; a bread_0; a lettuce_0; a spoon_0; a stoveburner_0; a pot_0; a stoveburner_1; a stoveburner_2; a stoveknob_0. On your right, you see a cabinet_17; a countertop_1; a cup_0; a plate_0; a butterknife_0; a potato_0; a saltshaker_0; a soapbottle_0; a tomato_0; a drawer_2; a stool_0. 
Assistant: Thought:  The sink_0 is not in my view. I need to turn to my right to find it.
Assistant: Act: turn_right.
System: Feedback: Action succeeded. Turned right by '90' degrees.
Obs: In front of you, You see a cabinet_0; a cabinet_9; a cabinet_11; a fridge_0. On your left, you see a cabinet_2; a cabinet_4; a cabinet_10; a cabinet_13; a countertop_2; a bowl_0; a faucet_1; a knife_0; a pan_0; a toaster_0; a diningtable_0; a apple_0; a bread_0; a butterknife_0; a lettuce_0; a potato_0; a saltshaker_0; a soapbottle_0; a spoon_0; a tomato_0; a drawer_3; a microwave_0; a sink_0; a sink_1; a stool_0; a window_0. On your right, you see a cabinet_3; a cabinet_5; a cabinet_12; a cabinet_21; a cabinet_22; a faucet_0. 
Assistant: Thought:  The sink_0 is in front of me. I need to move ahead and put the dishsponge back to the sink_0.
Assistant: Act: move_ahead.
System: Feedback: Action succeeded. Moved forward by 1 step.
Obs: In front of you, You see a cabinet_0; a cabinet_11; a fridge_0. On your left, you see a cabinet_1; a cabinet_2; a cabinet_4; a cabinet_6; a cabinet_9; a cabinet_10; a cabinet_18; a countertop_2; a bowl_0; a faucet_1; a knife_0; a pan_0; a toaster_0; a diningtable_0; a apple_0; a bread_0; a butterknife_0; a lettuce_0; a potato_0; a saltshaker_0; a soapbottle_0; a spoon_0; a tomato_0; a drawer_3; a stool_0; a window_0. On your right, you see a cabinet_3; a cabinet_5; a cabinet_12; a cabinet_13; a cabinet_14; a cabinet_21; a faucet_0; a microwave_0; a sink_0; a sink_1. 
Assistant: Thought:  I am in front of the sink_0. I need to drop the dishsponge_0 to put it back to its original position.
Assistant: Act: drop dishsponge_0 in sink_0.
System: Feedback: Action succeeded. You dropped dishsponge_0.
Inventory: Empty
Obs: In front of you, You see a cabinet_0; a cabinet_11; a fridge_0; a dishsponge_0. On your left, you see a cabinet_1; a cabinet_2; a cabinet_4; a cabinet_6; a cabinet_9; a cabinet_10; a cabinet_18; a countertop_2; a bowl_0; a faucet_1; a knife_0; a pan_0; a toaster_0; a diningtable_0; a apple_0; a bread_0; a butterknife_0; a lettuce_0; a potato_0; a saltshaker_0; a soapbottle_0; a spoon_0; a tomato_0; a drawer_3; a stool_0; a window_0. On your right, you see a cabinet_3; a cabinet_5; a cabinet_12; a cabinet_13; a cabinet_14; a cabinet_21; a faucet_0; a microwave_0; a sink_0; a sink_1. 
Assistant: Thought:  I have successfully reset the changed objects to their original states. Now, there is no change in the room. I can stop here.
Assistant: Act: stop []
System: You stopped the game.
System: {'misplaced': 0.0, 'fixed': 1.0}
System: [SUCCESS] You have completed the task. Congratulations!
System: DONE!
\end{lstlisting}

\subsection{Interactive QA}
\begin{lstlisting}
System: Imagine you are a robot that can move and see and open objects in a 2D grid kitchen environment.
 At the begining of this task I will give you a question about the environment and all the objects in front of you within your view size: 
 1. Question: You need to get the information of the environment and answer the question.
 2. Obs: Visible objects list 
 At each step you get some information from me, and you respond one robot thought and one robot action each time, nothing else, and then wait for feedback.
 You will get these information from me, remember these information are all from me, you can only generate Robot action: 
 1.Obs: Visible objects list
 2.Feedback: True or False, indicating the action is successful or not and why not.
 What you can do: 
 wait for me to give you next round feedback including Scene and Successful action and something else.
 Remember carefully, after you output robot thought and robot action, you should wait for me to give you feedback.
 Robot thought: 
 thought [thought info]: it means your thought about how to get the answer. 
 Robot action: 
 move_ahead: it means robot move ahead by 1 step.
 turn_left: it means robot turn left 90 degrees.
 turn_right: it means robot turn right 90 degrees.
 open [object_name]: it means robot open an openable object,open [fridge_0] means open fridge_0, and remerber do not issue action open if you have opened the object.
 answer [answer_info]: do this to answer the question. 
 Generate the thought and open and answer action in the correct format. Always put the action like thought [Found it] or open [fridge_0] or answer [True].
 Remember carefully,you generate thought and action,then wait for my feedback.And there is a space after the action name.

 Now i will give you 3 examples: 

 1.Question: Is there a mug in the room?
 Obs: In front of you, You see a Basketball_1,a Spoon_1,a Stool_1;On your left, you see a baseball_1.
 >thought [I should try my best to find the mug in the room.]
>move_ahead
Feedback: Action succeeded.
Obs: In front of you, You see a Basketball_2, a Spoon_1,a Plate_1.
>thought [Keep searching the room.]
>turn_right
Feedback: Action succeeded.
Obs: In front of you, You see a Basketball_3, a Pot_1, a fridge_1.

Feedback: Action succeeded.
Obs: In front of you, you see a mug_1, a Pot_1, a fridge_1.
>thought [Found it!]
>answer [True]

2.Question: I think there is a potato in the fridge. Is that correct?
Obs: In front of you, You see a potato_0,a winebottle_0,a cabinet_2;On your left, you see a baseball_1.
>thought [I should try my best to find the potato in the room.]
>move_ahead
Feedback: Action succeeded.
Obs: In front of you, You see a potato_0;a winebottle_0;a cabinet_2;On your left, you see a fridge_0.
>thought [Keep searching the room.]
>turn_right
Feedback: Action succeeded.
Obs: In front of you, You see a fridge_0.
>thought [Maybe in the fridge?]
>open [fridge_0]
Feedback: Action succeeded.
Obs: In/on it you You see mug_1,potato_0 in/on it.
>thought [Found it!]
>answer [True]

3.Question: Count the number of apples in this room.
Obs: In front of you, You see a bowl_0; a stool_0; a tomato_0; On your left, you see a butterknife_0; a cabinet_6; a coffeemachine_0; a book_0;
>thought [I should try my best to find all the apples in the room.]
>move_ahead
Feedback: Action succeeded.
Obs: In front of you, You see a bowl_0; a stool_0; On your left, you see a butterknife_0; a cabinet_6; a coffeemachine_0;
>thought [zero, keep looking.]
>move_ahead
Feedback: Action failed.
Obs: In front of you, You see a bowl_0; a stool_0; On your left, you see a butterknife_0; a cabinet_6; a coffeemachine_0;
>thought [zero, keep looking.]
>turn_right
Feedback: Action succeeded.
Obs: In front of you, You see a butterknife_0; a cabinet_6; a coffeemachine_0;
>thought [zero, keep looking.]
>move_ahead
Feedback: Action succeeded.
Obs: In front of you, You see a apple_0; a potato_1;a fridge_0; On your left, you see a cabinet_2;
>thought [One, keep looking.]
>open [fridge_0]
Feedback: Action succeeded.
Obs: In/on it you see apple_0; potato_1; apple_1; On your left, you see a cabinet_2;
>thought [Two, keep looking.]
>move_ahead
Feedback: Action succeeded.
Obs: In front of you, You see a potato_1; On your left, you see a stool_2;
>thought [Two, keep looking.]
>move_ahead
Feedback: Action succeeded.
Obs: In front of you, You see a apple_3; On your left, you see a stool_2;
>thought [Three!]
>answer [3]

Remember carefully,you need to generate noth thought and action!Then wait for my feedback,for example,what you nedd to generate is: thought [Maybe in the fridge?]
open [fridge_0].What i give you is : Feedback: Action succeeded.
Obs: In/on it you You see mug_1,potato_0 in/on it.
Respond YES if you can play this game.
Assistant: YES
System: Question: Do we have any cups?
Obs: You see nothing. You can try to take action like move_ahead, turn_left or turn_right to explore the room.  
Assistant: Act: move_ahead
System: Feedback: Action succeeded. Moved forward by '0.25' meter(s).
Obs: You see nothing. You can try to take action like move_ahead, turn_left or turn_right to explore the room.
Assistant: move_ahead
System: Feedback: Action succeeded. Moved forward by '0.25' meter(s).
Obs: You see nothing. You can try to take action like move_ahead, turn_left or turn_right to explore the room.
Assistant: Act: move_ahead
System: Feedback: Action succeeded. Moved forward by '0.25' meter(s).
Obs: You see nothing. You can try to take action like move_ahead, turn_left or turn_right to explore the room.
Assistant: Act: turn_right
System: Feedback: Action succeeded. Turn right by '90' degrees.
Obs: On your left, you see a lightswitch_0; On your right, you see a bottle_1, a creditcard_1, a mug_1, a creditcard_1, a stool_1, 
Assistant: Act: move_ahead
System: Feedback: Action succeeded. Moved forward by '0.25' meter(s).
Obs: On your left, you see a lightswitch_0; On your right, you see a bottle_1, a creditcard_1, a stool_1, 
Assistant: Act: turn_right
System: Feedback: Action succeeded. Turn right by '90' degrees.
Obs: In front of you, You see a stoveknob_0; a stoveknob_1; a stoveknob_2; a cabinet_5; a stoveknob_3; a houseplant_1; a spoon_1; On your left, you see a bowl_5; a drawer_1; a drawer_2; a cabinet_0; a cabinet_1; a window_0; a sink_0; a drawer_6; a cabinet_6; a drawer_8; a toaster_1; a winebottle_1; a vase_3; a coffeemachine_1; a bowl_5; a lettuce_1; a stool_1; a faucet_0; On your right, you see a bowl_3, a bottle_1, a bottle_1, a knife_1, a soapbottle_2, a vase_1, a bowl_3, a creditcard_1, a pot_1, a papertowelroll_1, a fridge_0, a lightswitch_0, a mug_1, a creditcard_1, 
Assistant: Thought: Found it! 
Assistant: Act: answer [True]
System: Your are right!
System: DONE!
\end{lstlisting}

\subsection{Household}
\begin{lstlisting}
System: You are an autonomous intelligent agent tasked with navigating a vitual home. You will be given a household task. These tasks will be accomplished through the use of specific actions you can issue.

Here's the information you'll have: 
Task: This is the task you are trying to accomplish.
Obs: These are the objects you can see.
Inventory: These are the objects you are holding.
Feedback: Whether the action is succeed or not and why is it.

At each step, you can choose to think your plan or execute an action from one of the following: 
`move_ahead`: it means you move ahead by 1 step.
`turn_left`: it means you turn left 90 degrees.
`turn_right`: it means you turn right 90 degrees.
`pick_up [object_name]`: Use this action to pick_up a pick_upable and visible object within your manipulation distance.
`drop [object_name]`: it means you places the object in hand at the place just in front of it.
`open [object_name]`: it means you open an openable and visible object within your manipulation distance.
`close [object_name]`: it means you close a closeable and visible object.
`toggle_on [object_name]`: it means you toggle a toggleable and visible object on.
`toggle_off [object_name]`: it means you toggle a toggleable and visible object off.
`slice [object_name]`: it means you slice a sliceable and visible object.
`put [object_name, receptacle_name]`: it means you places the object in hand into/onto the receptacle.
`stop [answer]`: Issue this action when you believe the task is complete. If you believe the task is impossible to complete, provide the answer as "N/A" in the bracket.

To be successful, it is very important to follow the following rules: 
1. You should only issue ONE action or ONE thought that is valid given the current observation
2. When you issue a action, and ONLY ONE action is enough, and then wait for your feedback.
3. Generate the action in the correct format starting with "Act: ". For example, "Act: pick_up [cup_0]".
4. State your thought or think process starting with "Thought: "
5. You can ONLY operate the object within your view distance.
6. If got action failed feedback, try to take other actions to proceed the task. 
7. Issue stop action when you think you have achieved the objective. Don't generate anything after stop.

Here are some examples: 
Task: put a clean lettuce in diningtable.
Obs: You can see nothing ahead.
> Thought: To solve the task, I need to find and take a lettuce, then clean it with sinkbasin, then put it in diningtable.
OK.
> Act: move_ahead

Respond YES if you can play this game.
Alfred Agent: YES
System: Task: Examine a phone by the light of the lamp in the corner. 
Obs: In front of you, You see a cellphone_0. On your left, you see a houseplant_0. On your right, you see a drawer_2; a sidetable_0; a lamp_2; a shelf_6; a shelf_7; a shelf_9; a shelf_11. 
What is your next step? Let's think before act.
Assistant: Thought:  To complete the task, I need to turn on the lamp, then pick up the phone to examine it. The lamp is within sight, so I can start by turning it on.
System: OK.
Assistant: Act: toggle_on [lamp_2]
System: Feedback: Action succeeded. You toggled lamp_2 on.
Obs: In front of you, You see a cellphone_0. On your left, you see a houseplant_0. On your right, you see a drawer_2; a sidetable_0; a lamp_2; a shelf_6; a shelf_7; a shelf_9; a shelf_11. 
Assistant: Act: pick_up [cellphone_0]
System: Feedback: Action succeeded. You picked cellphone_0 up.
Inventory: cellphone_0.
Obs: On your left, you see a houseplant_0. On your right, you see a drawer_2; a sidetable_0; a lamp_2; a shelf_6; a shelf_7; a shelf_9; a shelf_11. 
Assistant: Act: pick_up [cellphone_0]
System: Feedback: Action failed. There is no object "CellPhone|-05.53|+01.04|+00.17" existing. Please operate the object in sight.
Obs: On your left, you see a houseplant_0. On your right, you see a drawer_2; a sidetable_0; a lamp_2; a shelf_6; a shelf_7; a shelf_9; a shelf_11. 
Assistant: Thought:  I have picked up the cellphone and it is in my inventory. I'm now examining it by the light of the lamp. The task is complete.
System: OK.
Assistant: Act: stop [examined the phone by the light of the lamp]
System: You stopped the game.
System: 1.0
System: [SUCCESS] You have completed the task. Congratulations!
System: DONE!
\end{lstlisting}

\subsection{Multi-agent Teach}
\begin{lstlisting}
System: You are commander, an autonomous intelligent agent tasked with navigating a virtual home. You will be given a household task. These tasks will be accomplished through specific actions you post and through collaboration with follower. Note that you cannot interact with objects. You can only view the scene and guide follower to complete the task by using `chat`. You can not know the action history of follower.

Here's the information you'll have: 
Task: This is the task you are trying to accomplish.
Obs: These are the objects you can see. You can see at most 40.0 step(s) in front of you; 60 degrees on your left, 60 degrees on your right.
Inventory: These are the objects you are holding. You can hold at most 0 objects.
Feedback: Whether the action is succeed or not and why is it.

At each step, you can choose to think your plan or execute an action from one of the following: 
`move_ahead`: it means you move ahead by 1 step.
`move_back`: it means you move back by 1 step.
`turn_left`: it means you turn left 90 degrees.
`turn_right`: it means you turn right 90 degrees.
`no_op`: it means you do nothing.
`pan_left`: it means you move left by 1 step.
`pan_right`: it means you move right by 1 step.
`open_progress_check`: it means you check the progress of the task, you will be told which condition is failed.
`chat [chat_message]`: it means you send a chat to follower to guide it to do the task.
`select_oid [object_name]`: it means that you can select one object and get its position relative to the follower.
`search_object [object_type]`: it means that you can search for an object type and select one, which will then get its position relative to the follower.
`go_to [object_name]`: it means you travel to an available location next to the object.
`stop [answer]`: Issue this action when you believe the task is complete. If you believe the task is impossible to complete, provide the answer as "N/A" in the bracket.

To be successful, it is very important to follow the following rules: 
1. You should only issue an action or your thought that is valid given the current observation
2. You should only issue one action at a time.
3. Generate the action in the correct format. For example, pick_up [cup].
4. State your thought or think process starting with "Thought: "
5. Issue stop action when you think you have achieved the objective. Don't generate anything after stop.
6. Please do not generate the same operation as the last one.

Here are some examples: 
Task: put a clean lettuce on diningtable.
Obs: You can see nothing ahead.
> Thought: I need to check the task progress.
> Act: open_progress_check
Feedback: Action failed. One clean lettuce needs to be on dingingtable.
> Thought: I need to tell the task to follower.
> Act: chat [Please put a clean lettuce in diningtable.] 
Feedback: Action succeeded.
follower: Where can I find lettuce?
Thought: To solve the task, I need to guide follower to find and take a lettuce, then clean it with sinkbasin, then put it in diningtable.
> Act: select_oid [lettuce_0]
Feedback: Action succeeded. lettuce_0 is in front and left of follower.

Respond YES. if you can play this game.
System: You are follower, an autonomous intelligent agent tasked with navigating a virtual home. You will be given a household task. These tasks will be accomplished through specific actions you post and through collaboration with commander. Note that you can interact with objects but have no access to task information. You can use `chat` to ask commander for information to accomplish the task. You can not know the action history of follower.

Here's the information you'll have: 
Task: This is the task you are trying to accomplish.
Obs: These are the objects you can see. You can see at most 8.0 step(s) in front of you; 60 degrees on your left, 60 degrees on your right.
Inventory: These are the objects you are holding. You can hold at most 1 objects.
Feedback: Whether the action is succeed or not and why is it.

At each step, you can choose to think your plan or execute an action from one of the following: 
`move_ahead`: it means you move ahead by 1 step.
`move_back`: it means you move back by 1 step.
`turn_left`: it means you turn left 90 degrees.
`turn_right`: it means you turn right 90 degrees.
`no_op`: it means you do nothing.
`pan_left`: it means you move left by 1 step.
`pan_right`: it means you move right by 1 step.
`open_progress_check`: it means you check the progress of the task, you will be told which condition is failed.
`chat [chat_message]`: it means you send a chat to ask for information from commander to complete the task.
`go_to [object_name]`: it means you travel to an available location next to the object.
`pick_up [object_name]`: Use this action to pick_up a pickupable and visible object within 8.0 step(s).
`place [receptacle_name]`: it means you places the object in hand at the receptacle.
`open [object_name]`: it means you open an openable and visible object within 8.0 step(s).
`close [object_name]`: it means you close a closeable and visible object within 1 step ahead.
`toggle_on [object_name]`: it means you toggle a toggleable and visible object on within 1 step ahead.
`toggle_off [object_name]`: it means you toggle a toggleable and visible object off within 1 step ahead.
`slice [object_name]`: it means you slice a sliceable and visible object within 1 step ahead.
`stop [answer]`: Issue this action when you believe the task is complete. If you believe the task is impossible to complete, provide the answer as "N/A" in the bracket.

To be successful, it is very important to follow the following rules: 
1. You should only issue an action or your thought that is valid given the current observation
2. You should only issue one action at a time.
3. Generate the action in the correct format. For example, pick_up [cup].
4. State your thought or think process starting with "Thought: "
5. Issue stop action when you think you have achieved the objective. Don't generate anything after stop.
6. Please do not generate the same operation as the last one.

Here are some examples: 
Task: You can use `chat` to ask commander for task information.
Obs: You can see nothing ahead.
> Thought: I need to wait information from commander.
> Act: no_op
Feedback: Action succeeded.
commander: Please put a clean lettuce in diningtable.
> Thought: I need to ask commander for the location of the lettuce.
> Act: chat [Where can I find lettuce?]
Feedback: Action succeeded.
Obs: You can see nothing ahead.
> Thought: To solve the task, I need to find and take a lettuce, then clean it with sinkbasin, then put it in diningtable.
But I don't know where lettuce is, I can wait information from follower or explore by myself.
> Act: pan_left
Feedback: Action succeeded.
Obs: In front of you, You see a lettuce_0.

Respond YES. if you can play this game.
commander: YES.
System: Task: Put all Pillow on any ArmChair.
Obs: In front of you, You see a lightswitch_0; a closed drawer_0; a sidetable_0, there is a desklamp_1, a creditcard_0, a remotecontrol_0, on it. On your left, you see a shelf_0; a shelf_1; a shelf_2, there is a desklamp_1, on it; a shelf_3. On your right, you see a garbagecan_0; a painting_0; a dresser_0, there is a pillow_0, a desklamp_2, a newspaper_1, a keychain_1, a cellphone_0, an opened laptop_0, it's empty, a vase_0, on it; a closed drawer_3; a closed cabinet_3. 
What is your next step? Let's think and act.
follower: YES.
System: Task: You can use `chat` to ask commander for task information.
Obs: On your left, you see a dresser_0, there is a pillow_0, a desklamp_2, a newspaper_1, a keychain_1, a cellphone_0, an opened laptop_0, it's empty, a vase_0, on it. On your right, you see a closed drawer_1; a closed cabinet_0. 
What is your next step? Let's think and act.
commander: Thought:  I need to find an ArmChair to place the pillows on. But first, I'll ask the follower to pick up the pillow from the dresser.
commander: Act: chat [Please pick up the pillow from the dresser.]
System: Feedback: Action succeeded.
follower: Thought:  The pillow is visible on the dresser. I need to go to the dresser and pick up the pillow.
follower: Act: go_to [dresser_0]
System: Feedback: Action succeeded. Go to dresser_0. In front of you, You see a garbagecan_0; a closed drawer_2; a closed cabinet_1. On your left, you see a dresser_0, there is a pillow_0, a desklamp_2, a newspaper_1, a keychain_1, a cellphone_0, an opened laptop_0, it's empty, a vase_0, on it; a closed drawer_3; a closed cabinet_3. On your right, you see a closed drawer_1; a closed cabinet_0. 
commander: Thought:  Next, I need to guide the follower to an armchair to put the pillow on it. I should find an armchair first.
commander: Act: search_object [ArmChair]
System: Feedback: Action succeeded. armchair_0 is at the left rear of follower.
follower: Act: pick_up [pillow_0]
System: Feedback: Action succeeded. You picked pillow_0 up.
Inventory: pillow_0.
Obs: In front of you, You see a garbagecan_0; a closed drawer_2; a closed cabinet_1. On your left, you see a dresser_0, there is a pillow_0, a desklamp_2, a newspaper_1, a keychain_1, a cellphone_0, an opened laptop_0, it's empty, a vase_0, on it; a closed drawer_3; a closed cabinet_3. On your right, you see a closed drawer_1; a closed cabinet_0. 
commander: Act: chat [please place the pillow on the armchair to your left rear.]
System: Feedback: Action succeeded.
follower: Thought:  Before placing the pillow on the armchair, I should ensure the armchair is visible. I will turn left twice to make the armchair visible.
follower: Act: turn_left
System: Feedback: Action succeeded. Turn left by '90' degrees.
Obs: You see nothing. You can try to take action like move_ahead, turn_left or turn_right to explore the room.
commander: Act: search_object [pillow]
System: Feedback: Action succeeded. pillow_0 is in front and left of follower.
follower: Act: turn_left
System: Feedback: Action succeeded. Turn left by '90' degrees.
Obs: In front of you, You see a armchair_0, there is a keychain_0, on it; a closed drawer_2; a closed cabinet_1; a closed drawer_4. On your left, you see a closed cabinet_2; a dresser_0, there is a pillow_0, a desklamp_2, a newspaper_1, a keychain_1, a cellphone_0, an opened laptop_0, it's empty, a vase_0, on it. 
commander: Act: chat [please pick up the pillow in front of you.]
System: Feedback: Action succeeded.
follower: Act: go_to [armchair_0]
System: Feedback: Action succeeded. Go to armchair_0. In front of you, You see a armchair_0, there is a keychain_0, on it; a closed drawer_2; a closed cabinet_1; a closed drawer_4. On your left, you see a closed cabinet_2; a dresser_0, there is a pillow_0, a desklamp_2, a newspaper_1, a keychain_1, a cellphone_0, an opened laptop_0, it's empty, a vase_0, on it. 
commander: Act: chat [please place the pillow on the same armchair.]
System: Feedback: Action succeeded.
follower: Act: place [armchair_0]
System: Feedback: Action succeeded. You put pillow_0 to armchair_0.
Obs: In front of you, You see a armchair_0, there is a keychain_0, a pillow_0, on it; a closed drawer_2; a closed cabinet_1; a closed drawer_4. On your left, you see a closed cabinet_2; a dresser_0, there is a pillow_0, a desklamp_2, a newspaper_1, a keychain_1, a cellphone_0, an opened laptop_0, it's empty, a vase_0, on it. 
System: [SUCCESS] You have completed the task. Congratulations!
System: DONE!
\end{lstlisting}

\subsection{Multi-agent Cooperation}
\begin{lstlisting}
System: You are an autonomous intelligent agent tasked with navigating a virtual home. Your name is Alice. You are hurry to finish a housework with your friend Bob together. There are four kinds of rooms, livingroom, kitchen, bedroom, bathroom. You will be given a household task. These tasks will be accomplished through the use of specific actions you and your friend can issue. You have two hands to pickup, hold, put and check objects.

Here's the information you will have: 
Task: This is the task you and your frend Bob are trying to accomplish.
At each step, you can choose to think your plan or issue an action from one of the following: 
`chat [message].`: You can use this action to send a message to your friend Bob.
`go_explore [room].`: Use this action to go to a room and explore around.
`go_check [container].`: Use this action to go to a unchecked container and check its inside. You must have at least one free hands to check.
`go_grab [object]`: Use this action to go to an object and grab it.
`go_put [container].`: Use this action to go to the container and place the object in the container. You need to make sure you have the object(s) you need for the task.
`stop [answer].`: Issue this action when you believe the task is complete. If you believe the task is impossible to complete, provide the answer as "N/A" in the bracket.

To be successful, it is very important to follow the following rules: 
1. You should only issue an action or your thought that is valid given the current observation
2. You should only issue one action at a time.
3. Generate the action in the correct format starting with "Act: ". For example, "Act: go_grab [phone_0].".
4. State your thought or think process starting with "Thought: "
5. Issue stop action when you think you have achieved the objective. Don't generate anything after stop.
6. You should generate chats like this: chat [Hello, how can I help you?], taking care to enclose them in square brackets.

Respond YES. if you can play this game.
System: You are an autonomous intelligent agent tasked with navigating a virtual home. Your name is Bob. You are hurry to finish a housework with your friend Alice together. There are four kinds of rooms, livingroom, kitchen, bedroom, bathroom. You will be given a household task. These tasks will be accomplished through the use of specific actions you and your friend can issue. You have two hands to pickup, hold, put and check objects.

Here's the information you will have: 
Task: This is the task you and your frend Alice are trying to accomplish.
At each step, you can choose to think your plan or issue an action from one of the following: 
`chat [message].`: You can use this action to send a message to your friend Alice.
`go_explore [room].`: Use this action to go to a room and explore around.
`go_check [container].`: Use this action to go to a unchecked container and check its inside. You must have at least one free hands to check.
`go_grab [object]`: Use this action to go to an object and grab it.
`go_put [container].`: Use this action to go to the container and place the object in the container. You need to make sure you have the object(s) you need for the task.
`stop [answer].`: Issue this action when you believe the task is complete. If you believe the task is impossible to complete, provide the answer as "N/A" in the bracket.

To be successful, it is very important to follow the following rules: 
1. You should only issue an action or your thought that is valid given the current observation
2. You should only issue one action at a time.
3. Generate the action in the correct format starting with "Act: ". For example, "Act: go_grab [phone_0].".
4. State your thought or think process starting with "Thought: "
5. Issue stop action when you think you have achieved the objective. Don't generate anything after stop.
6. You should generate chats like this: chat [Hello, how can I help you?], taking care to enclose them in square brackets.

Respond YES. if you can play this game.
Alice: YES.
System: Task: Find and put 1 wine onto the coffeetable_0.
Obs: You have already found and put pudding_0, juice_0, juice_1 onto the coffeetable_0. You are holding nothing. You are in the kitchen, where you found unchecked containers kitchencabinet_0, kitchencabinet_1, kitchencabinet_2, kitchencabinet_3, kitchencabinet_4, kitchencabinet_5, kitchencabinet_6, kitchencabinet_7, stove_0, dishwasher_0, fridge_0, fridge_1, microwave_0. You don't know where Bob is. The livingroom is unexplored. The bedroom is unexplored. The bathroom is unexplored. 
What is your next step? Let's think and act.
Bob: YES.
System: Task: Find and put 1 wine onto the coffeetable_0.
Obs: You have already found and put pudding_0, juice_0, juice_1 onto the coffeetable_0. You are holding nothing. You are in the bedroom, where you found an unchecked container cabinet_0. You don't know where Alice is. The livingroom is unexplored. The kitchen is unexplored. The bathroom is unexplored. 
What is your next step? Let's think and act.
Alice: Thought:  The task is to find and put 1 wine onto the coffeetable_0. I am currently in the kitchen with several unchecked containers. I will start by checking these containers as the wine could be in any of them.
Alice: Act: go_check [kitchencabinet_0].
System: Feedback: Action succeeded. You opened kitchencabinet_0.
Obs: In it you see nothing
Bob: Thought:  Since I need to find the wine and I am currently in an unexplored bedroom, I'll first check the unchecked container in the bedroom, the cabinet_0. If the wine is not there, I'll move to the next unexplored room.
Bob: Act: go_check [cabinet_0].
System: Feedback: Action succeeded. You opened cabinet_0.
Obs: In it you see book_3
Alice: Thought:  The first kitchen cabinet was empty. I'll check the next one.
Alice: Act: go_check [kitchencabinet_1]
System: Feedback: Action succeeded. You opened kitchencabinet_1.
Obs: In it you see nothing
Bob: Thought:  The wine is not in the cabinet_0 in the bedroom, so I need to explore other rooms. The kitchen is a good place to start as it is common to store wine there.
Bob: Act: go_explore [kitchen]
System: Feedback: Action succeeded. Go to kitchen_0. You have already found and put pudding_0, juice_0, juice_1 onto the coffeetable_0. You are holding nothing. You are in the kitchen, where you found unchecked containers kitchencabinet_2, kitchencabinet_3, kitchencabinet_4, kitchencabinet_5, kitchencabinet_6, kitchencabinet_7, stove_0, dishwasher_0, fridge_0, fridge_1, microwave_0. You also see Alice here in the kitchen, she is holding nothing. The livingroom is unexplored. You found nothing in the bedroom. The bathroom is unexplored. 
Alice: Act: go_check [kitchencabinet_2]
System: Feedback: Action succeeded. You opened kitchencabinet_2.
Obs: In it you see book_1
Bob: Act: go_check [fridge_0]
System: Feedback: Action succeeded. You opened fridge_0.
Obs: In it you see wine_0
Alice: Act: go_check [kitchencabinet_3]
System: Feedback: Action succeeded. You opened kitchencabinet_3.
Obs: In it you see nothing
Bob: Act: go_grab [wine_0]
System: Feedback: Action succeeded. You picked wine_0 up.
Obs: You have already found and put pudding_0, juice_0, juice_1 onto the coffeetable_0. You are holding wine_0. You are in the kitchen, where you found unchecked containers kitchencabinet_4, kitchencabinet_5, kitchencabinet_6, kitchencabinet_7, stove_0, dishwasher_0, fridge_1, microwave_0. You also see Alice here in the kitchen, she is holding nothing. The livingroom is unexplored. You found nothing in the bedroom. The bathroom is unexplored. 
Alice: Act: go_check [kitchencabinet_4]
System: Feedback: Action succeeded. You opened kitchencabinet_4.
Obs: In it you see nothing
Bob: Act: go_explore [livingroom]
System: Feedback: Action succeeded. Go to livingroom_0. You have already found and put pudding_0, juice_0, juice_1 onto the coffeetable_0. You are holding wine_0. You are in the livingroom, where you found coffeetable_0. Last time you saw Alice was in the kitchen, she was holding nothing. You found unchecked containers kitchencabinet_5, kitchencabinet_6, kitchencabinet_7, stove_0, dishwasher_0, fridge_1, microwave_0 in the kitchen. You found nothing in the bedroom. The bathroom is unexplored. 
Alice: Act: go_check [kitchencabinet_5]
System: Feedback: Action succeeded. You opened kitchencabinet_5.
Obs: In it you see book_2
Bob: Act: go_put [coffeetable_0]
System: Feedback: Action succeeded. You put wine_0 on coffeetable_0.
Obs: You have already found and put pudding_0, juice_0, juice_1, wine_0 to the coffeetable_0. You are holding nothing. You are in the livingroom, where you found coffeetable_0. Last time you saw Alice was in the kitchen, she was holding nothing. You found unchecked containers kitchencabinet_5, kitchencabinet_6, kitchencabinet_7, stove_0, dishwasher_0, fridge_1, microwave_0 in the kitchen. You found nothing in the bedroom. The bathroom is unexplored. 
System: [SUCCESS] You have completed the task. Congratulations!
System: DONE!
\end{lstlisting}
